\documentclass[10pt,onecolumn,twocolumn,letterpaper]{article}

\usepackage{amsmath}
\usepackage{amssymb}
\usepackage{bm}
\usepackage{tikz}
\usetikzlibrary{arrows,3d,calc}
\usepackage{tikz-3dplot}
\usepackage{booktabs}
\usepackage{caption}
\usepackage{enumitem}
\usepackage{graphicx}
\usepackage{subcaption}
\usepackage{verbatim}
\usepackage[pagenumbers]{cvpr}
\usepackage[pagebackref,breaklinks,colorlinks]{hyperref}
\usepackage[capitalize]{cleveref}
\usepackage[colorinlistoftodos,textwidth=20mm]{todonotes}

\crefname{section}{Sec.}{Secs.}
\Crefname{section}{Section}{Sections}
\Crefname{table}{Table}{Tables}
\crefname{table}{Tab.}{Tabs.}

\begin{document}

\title{\textbf{Limitations of (Procrustes) Alignment in Assessing Multi-Person Human Pose and Shape Estimation}}

\author{
    Drazic Martin, Pierre Perrault\\
    \\
    Idemia
    \\
    {\tt \small \{drazic.martin, pierre.perrault\}
    @ \href{https://www.polytechnique.edu/}{idemia.com}}
}

\makeatletter

\def\@maketitle{
    \newpage \null \vskip 2em
    \begin{center}
        \let \footnote \thanks
        {\LARGE \@title \par} \vskip 3em
        {\large \lineskip .2em
            \begin{tabular}[t]{c}
                \@author
            \end{tabular} \par
        } \vskip 1em
    {\large \@date}
    \end{center} \par \vskip 1.5em
}

\maketitle

\begin{abstract}
We delve into the challenges of accurately estimating 3D human pose and shape in video surveillance scenarios. Beginning with the advocacy for metrics like W-MPJPE and W-PVE, which omit the (Procrustes) realignment step, to improve model evaluation, we then introduce RotAvat. This technique aims to enhance these metrics by refining the alignment of 3D meshes with the ground plane. Through qualitative comparisons, we demonstrate RotAvat's effectiveness in addressing the limitations of existing approaches.
\end{abstract}
\section{Introduction}
\label{sec:intro}

In today's rapidly evolving technological landscape, video surveillance has emerged as a crucial tool for maintaining security and enhancing situational awareness in a variety of contexts. As the ubiquity of surveillance cameras continues to grow, so does the need for advanced scene understanding algorithms, especially in scenarios involving pedestrians. %These algorithms play an increasingly vital role in applications ranging from public safety and traffic management to human-computer interaction and immersive virtual environments. 
The accurate estimation of pedestrian pose and shape in 3D
%, coupled with mesh reconstruction, 
is at the forefront of addressing these evolving demands, making it a focal point of contemporary research and development efforts.

In this short paper, we investigate scenarios that faithfully represent the aforementioned video surveillance situations. These scenarios involve a stationary camera, such as a CCTV, capturing either a single RGB image or a video stream in environments containing pedestrians. The assumption of a static camera is mild given its relevance to the real-world applications we are considering. 
Moreover, in our context, the multi-person aspect is crucial, particularly for public spaces like malls, railway stations and airports, where the simultaneous presence of many individuals is common.

The task under consideration 
is to determine the 3D human pose and shape (HPS) of every person within the frame. The HPS task has been thoroughly investigated in previous research studies. Nevertheless, we contend that the existing literature inadequately addresses the scenarios outlined above. Specifically, within a video surveillance framework, a comprehensive understanding of the global 3D scene is essential. This means that the 3D position of bodies in the world coordinate system holds equal or greater importance than their 3D pose and shape. However, we argue that the metrics presently employed (and, consequently, the current state-of-the-art approaches) tend to significantly undervalue the former aspect, prioritizing local pose and shape instead.

In light this, we advocate for the utilization of suitable metrics, such as the W-MPJPE and W-PVE metrics introduced by SPEC \cite{SPEC:ICCV:2021}. The main obstacle to a wider adoption of these metrics lies in the lack of performance of current models at the level of global coordinates. We therefore present a methodology aimed at effectively addressing this challenge faced by current HPS models in the video surveillance context. Our proposed approach requires no training and can be applied as a post-processing step to any existing solution. We finally assess its effectiveness qualitatively in comparison with the most recent methods.

\section{Related work}
Here, we give a concise review of the literature on HPS, emphasizing key studies in this domain in connection to the identified issues. The majority of current methods \cite{zhang2021pymaf,yi2023mime, tripathi20233d, black2023bedlam} utilize a 2-stages detection-based pipeline to infer HPS and lack crucial information such as camera calibration data. Yet, some recent 2-stages and 1-stage  methods have begun to incorporate full-image information into their approaches. For example, SPEC \cite{SPEC:ICCV:2021} estimates the perspective camera from a single full-image and employs it to reconstruct HPS in a 2-stages manner.
Another 2-stages approach, 
CLIFF \cite{li2022cliff}, uses information about the camera by concatenating the feature of the cropped image with its bounding box location.
ROMP \cite{ROMP}, on the other hand, directly estimates multiple individuals simultaneously from the entire image by predicting both a body center heatmap and a mesh parameter map, enabling a description of the 3D body mesh at the pixel level. BEV \cite{BEV} surpasses ROMP by introducing an imaginary bird’s-eye-view map. This map, when combined with the front-view maps, enables the construction of a 3D view in camera coordinates. 

\paragraph{Issues with existing methods}
Existing HPS methods still face significant difficulties in accurately predicting the global 3D position of humans, particularly when applied to \emph{in the wild} scenarios characterized by various conditions, such as crowds (e.g. hundreds of people), unusual camera positions and scale variations. The main difficulty stems from the scarcity of sufficient multi-person data with accurate 3D translation annotations for supervision in diverse environments with varying camera perspectives, necessary for purely deep learning-based HPS methods. Yet, even if such 3D annotated data were available, effectively utilizing it along with image inputs remains a challenge for deep learning methods, as 3D data generally goes beyond pixel-level information. Until such an approach can be trained to convincingly generalize to any scene, an alternative approach (which we advocate here) is the use of global 3D data in a \emph{deep-learning free} post-processing step.

\paragraph{Other recent related approaches} Other methods such that HybrIK \cite{li2021hybrik, li2023hybrik} integrates 3D keypoint estimation and body mesh recovery into a unified framework. There also exist methods, such as GLAMR \cite{yuan2022glamr} and TRACE \cite{TRACE}, that handle videos with dynamic camera cases.
\section{Metrics}
% Sect : the metrics (look at SPEC, they explain it well, also in their appendix)

The mean per joint position error (MPJPE), Procrustes-aligned mean per joint position error (PA-MPJPE), and per vertex error (PVE) stand out as widely employed evaluation metrics in literature.
In \cite{SPEC:ICCV:2021}, the authors raised concerns regarding the PA-MPJPE metric, highlighting its tendency to mask specific inaccuracies by removing body rotation, translation, or scaling effects. When compared to the Procrustes-aligned variants, MPJPE and PVE are commonly seen as stricter because they account for differences in rotation and scale. Yet, in real-world scenarios where camera viewpoints vary, MPJPE and PVE may not provide a clear assessment of method performance, as the predicted body and the ground truth body are still translated so that they share the same root joint. Thus, \cite{SPEC:ICCV:2021} introduced the W-MPJPE and W-PVE metrics, that compute the error in world coordinates, without any realignment step. These last metrics provide a fair benchmark of the whole HPS task, encompassing both 3D pose and shape, as well as 3D position in the world coordinate system.
\section{Our approach and how it compares with existing solutions}
% Sect : our approach ==> avoid the pure deep learning

\begin{figure}[h]
    \begin{subfigure}[h]{\linewidth}
        \includegraphics[width=0.495\linewidth]{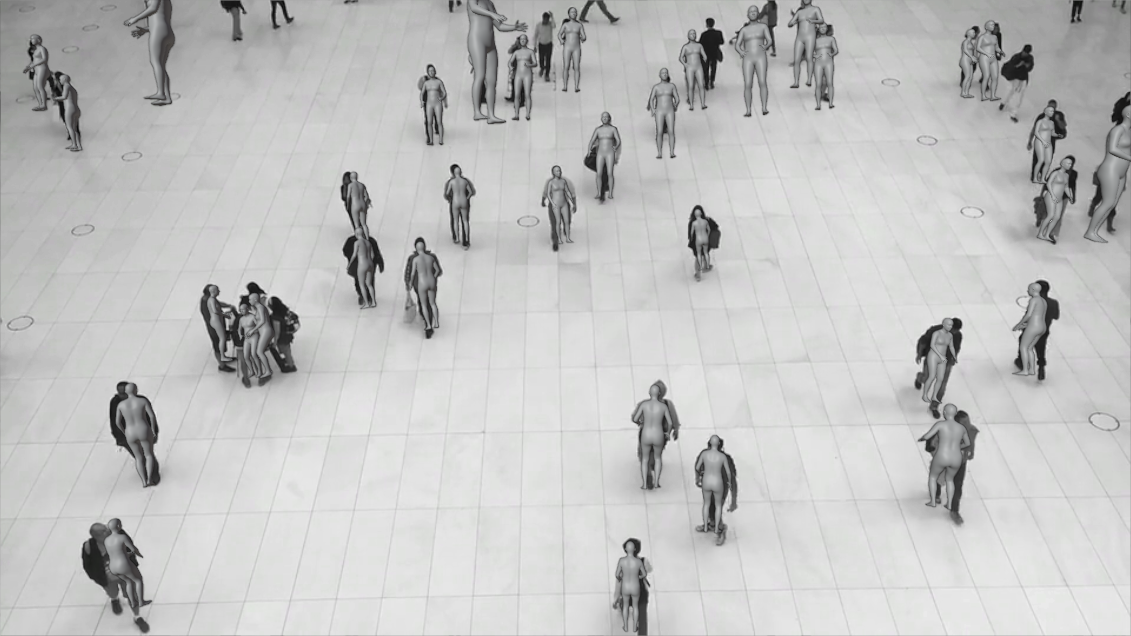}
        \hfill\includegraphics[width=.495\linewidth]{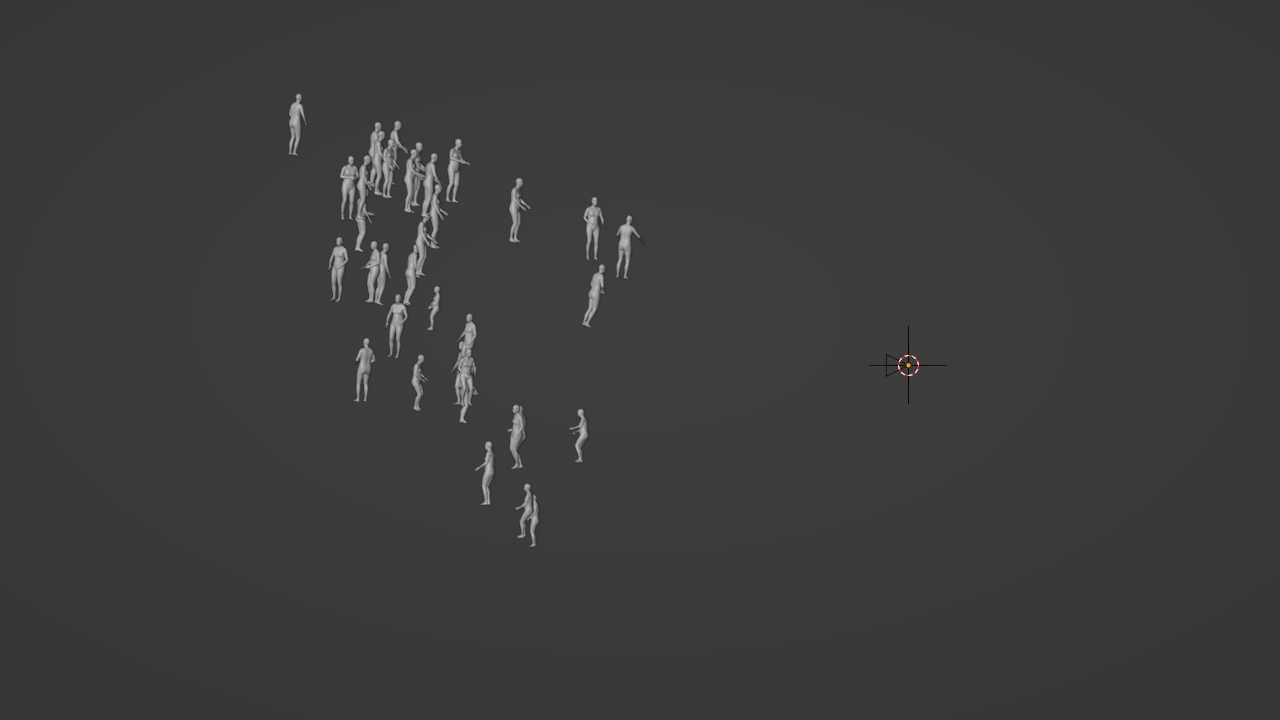}
            \caption{BEV \cite{BEV}
            }
    \end{subfigure}
    \begin{subfigure}[h]{\linewidth}
        \includegraphics[width=0.495\linewidth]{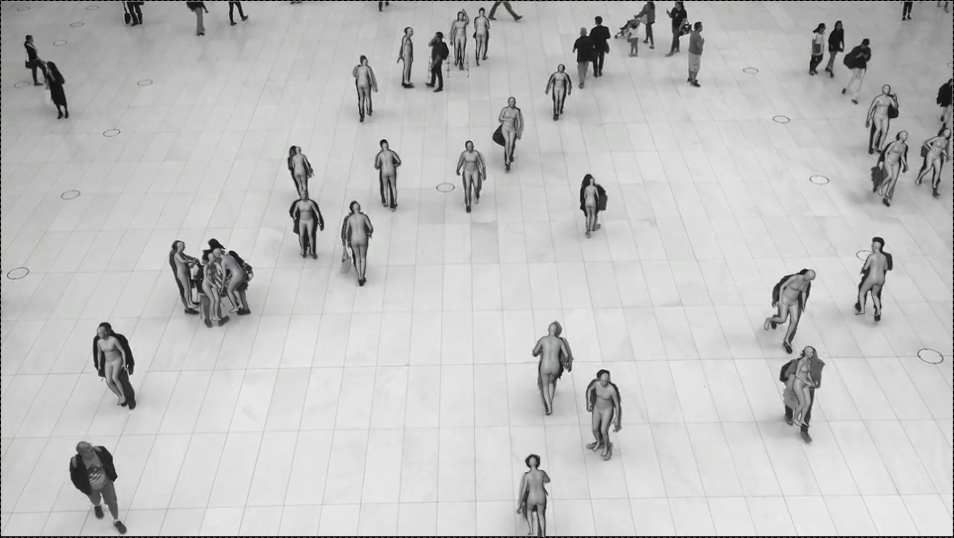}
        \hfill\includegraphics[width=.495\linewidth]{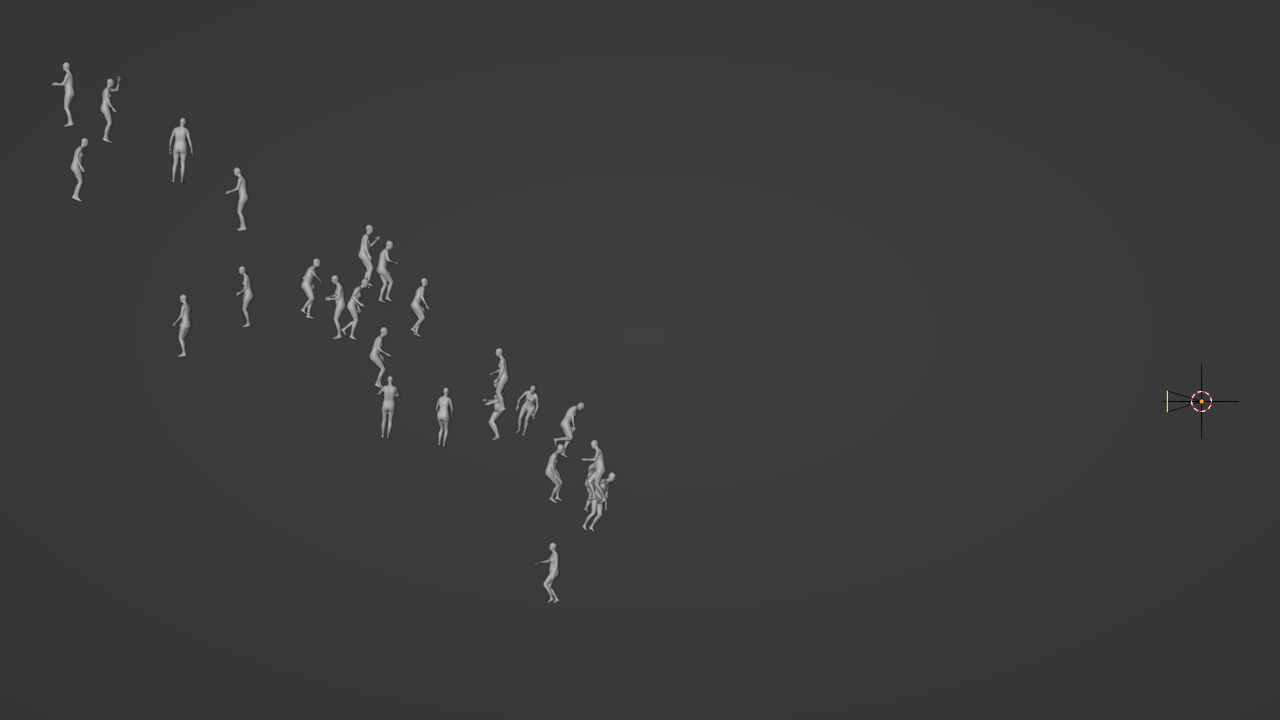}
            \caption{SPEC \cite{SPEC:ICCV:2021}}
    \end{subfigure}

    \begin{subfigure}[h]{\linewidth}
        \includegraphics[width=0.495\linewidth]{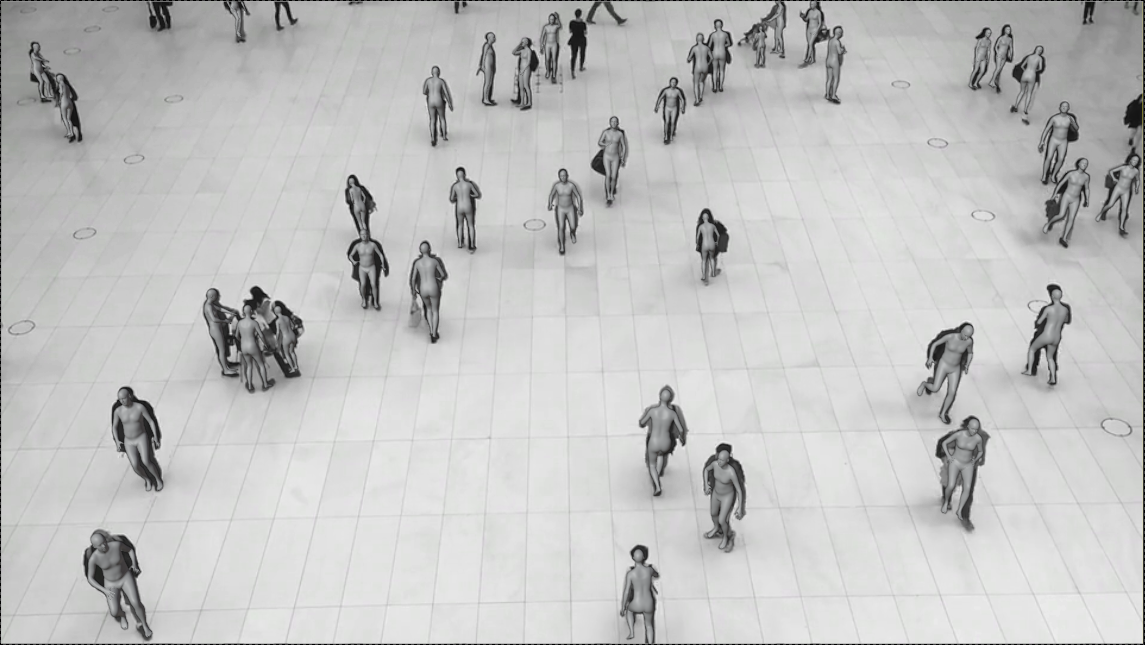}
        \hfill\includegraphics[width=.495\linewidth]{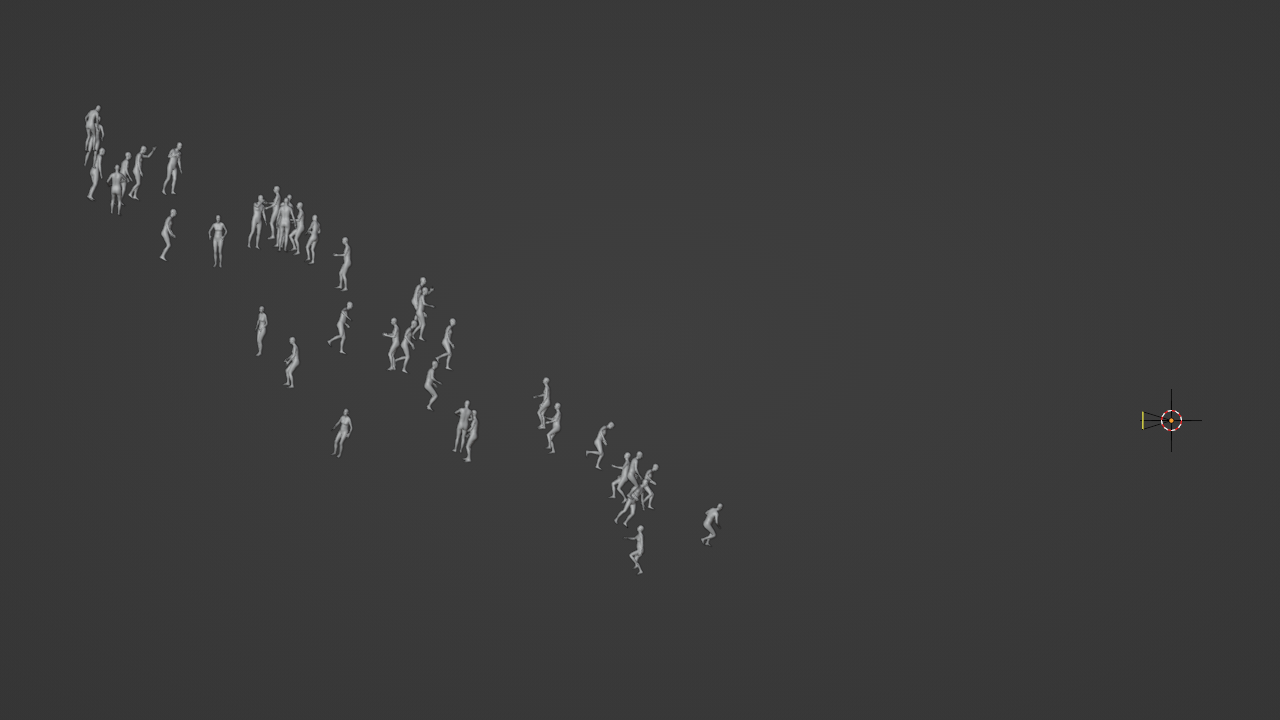}
            \caption{CLIFF \cite{li2022cliff}}
    \end{subfigure}

    \begin{subfigure}[h]{\linewidth}
        \includegraphics[width=0.495\linewidth]{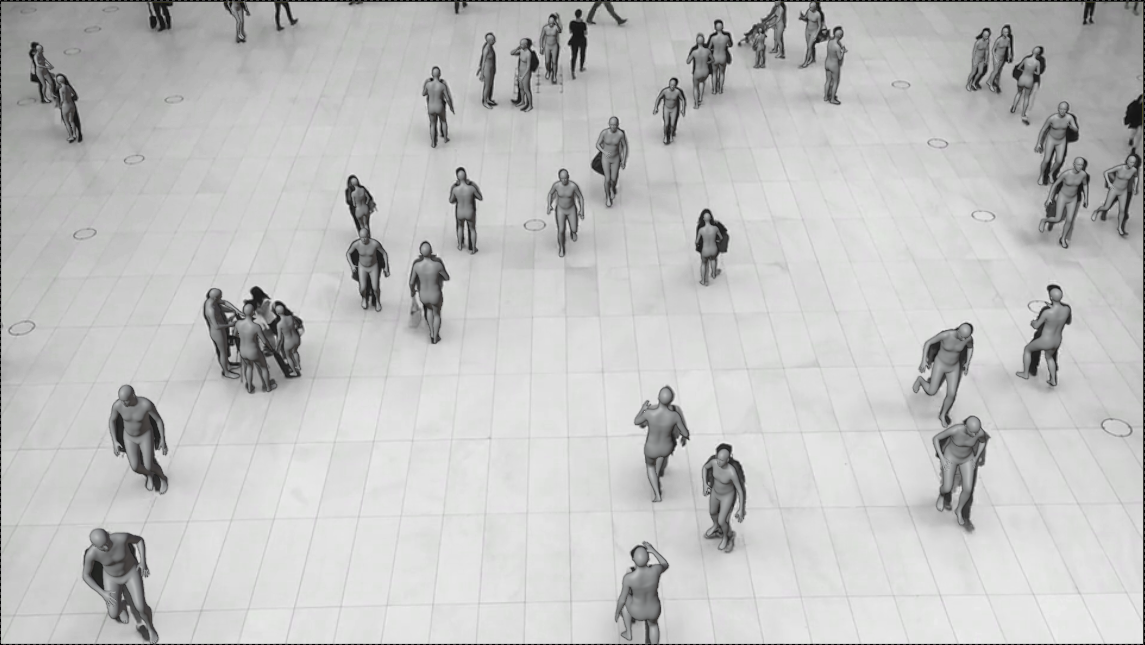}
        \hfill\includegraphics[width=.495\linewidth]{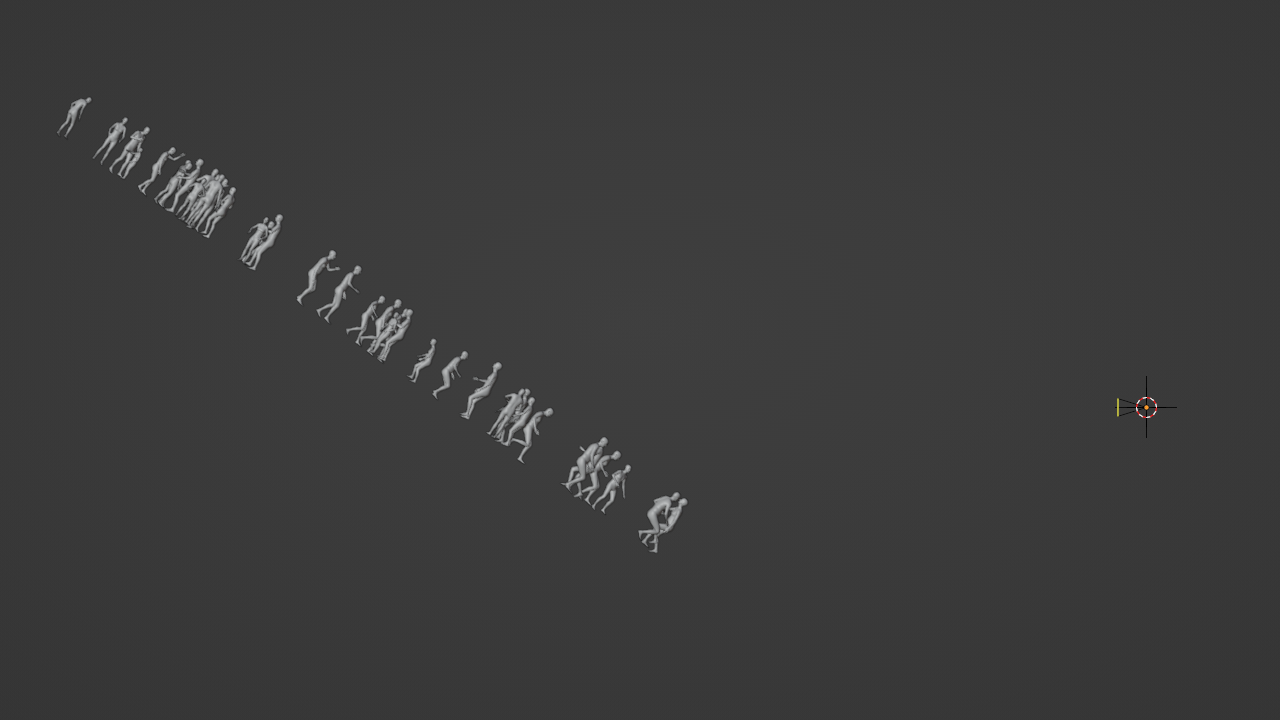}
            \caption{\textbf{RotAvat (ours)}}
    \end{subfigure}
    \caption{Comparison between BEV, SPEC, CLIFF and our proposed method, front and side views. Note that we rendered the side view with an orthographic projection to better appreciate meshes elevation relative to the ground.}
    \label{fig_crowd_indoor}
\end{figure}

\begin{figure}[h]
    \begin{subfigure}[h]{\linewidth}
        \includegraphics[width=0.495\linewidth]{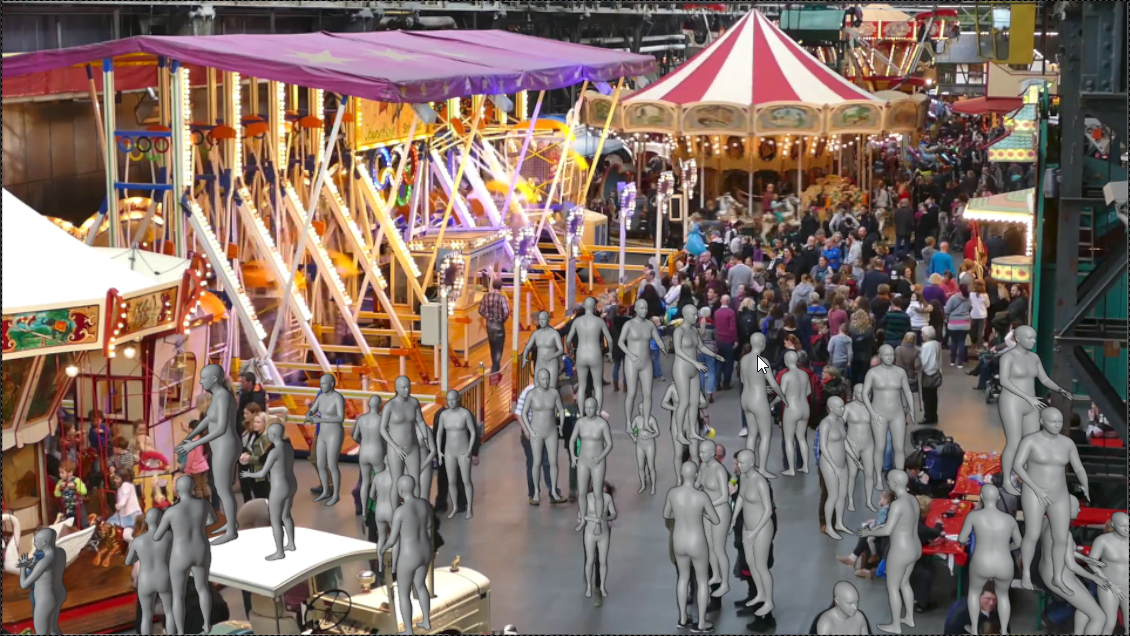}
        \hfill\includegraphics[width=.495\linewidth]{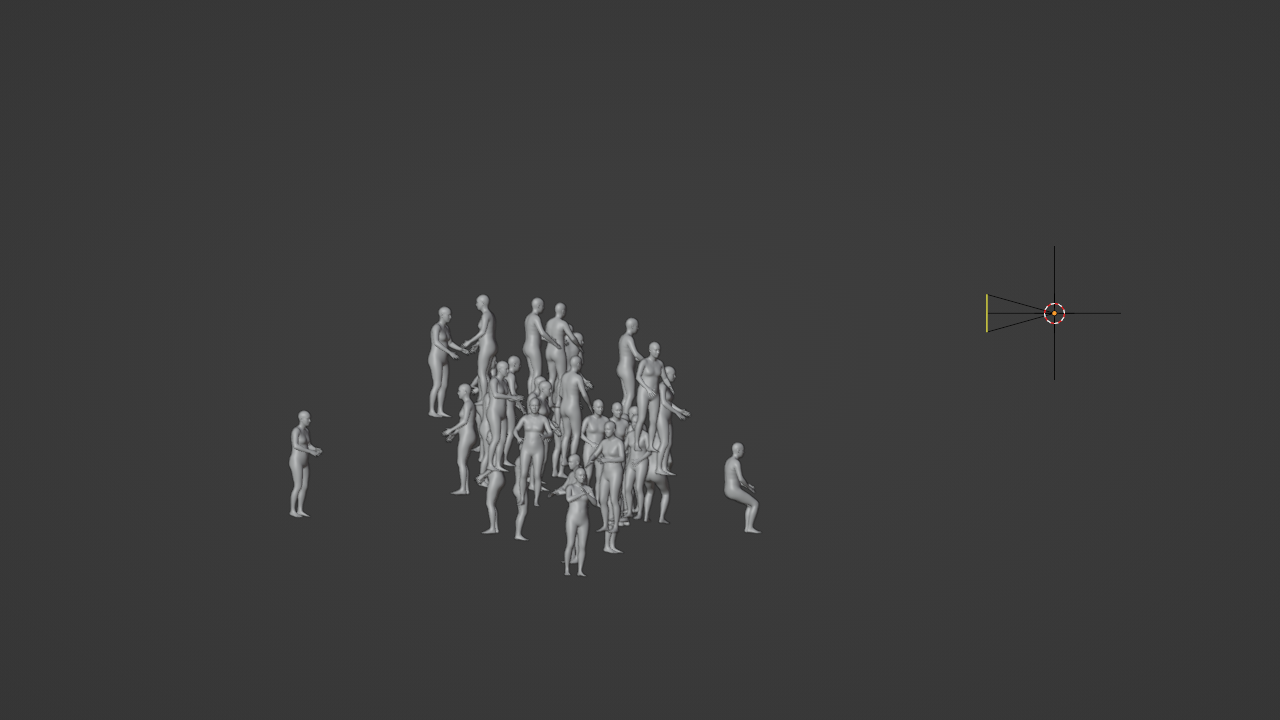}
            \caption{BEV \cite{BEV}
            }
    \end{subfigure}
    \begin{subfigure}[h]{\linewidth}
        \includegraphics[width=0.495\linewidth]{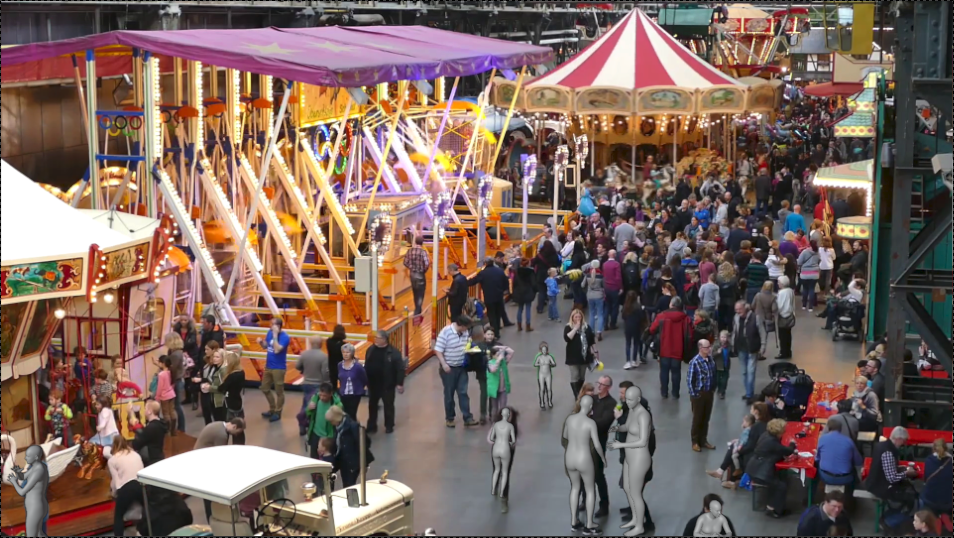}
        \hfill\includegraphics[width=.495\linewidth]{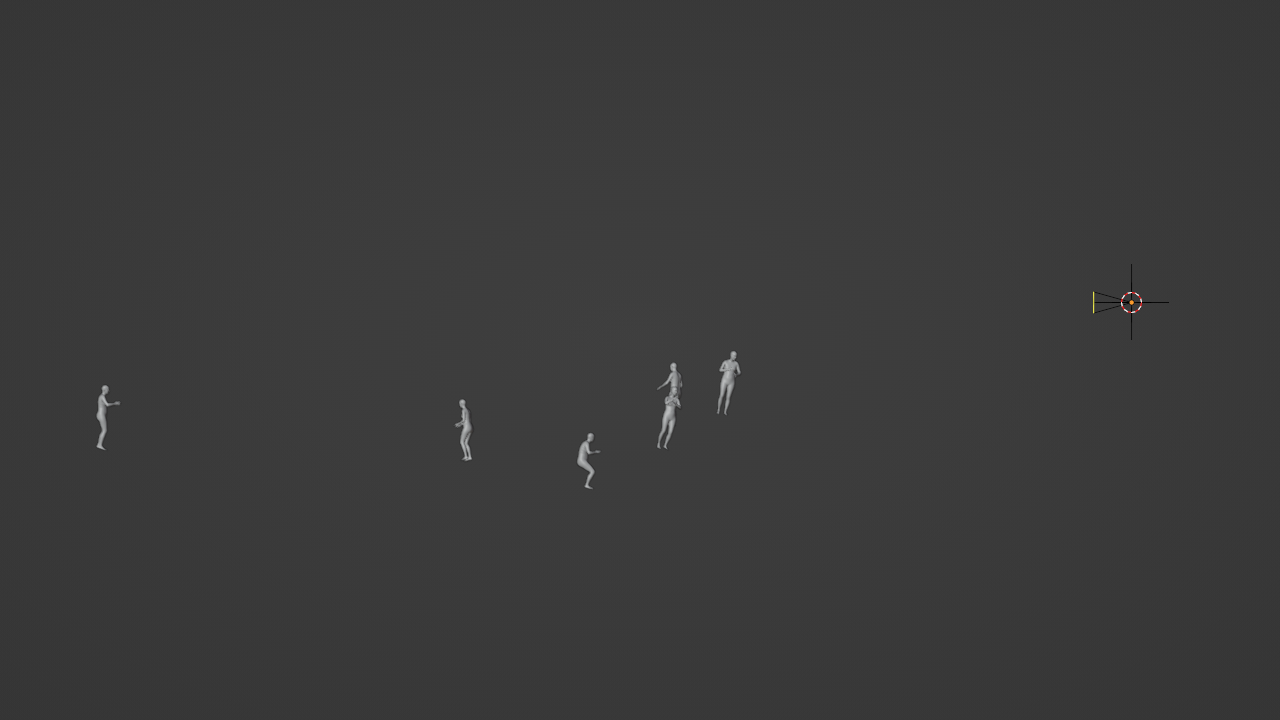}
            \caption{SPEC \cite{SPEC:ICCV:2021}}
    \end{subfigure}

    \begin{subfigure}[h]{\linewidth}
        \includegraphics[width=0.495\linewidth]{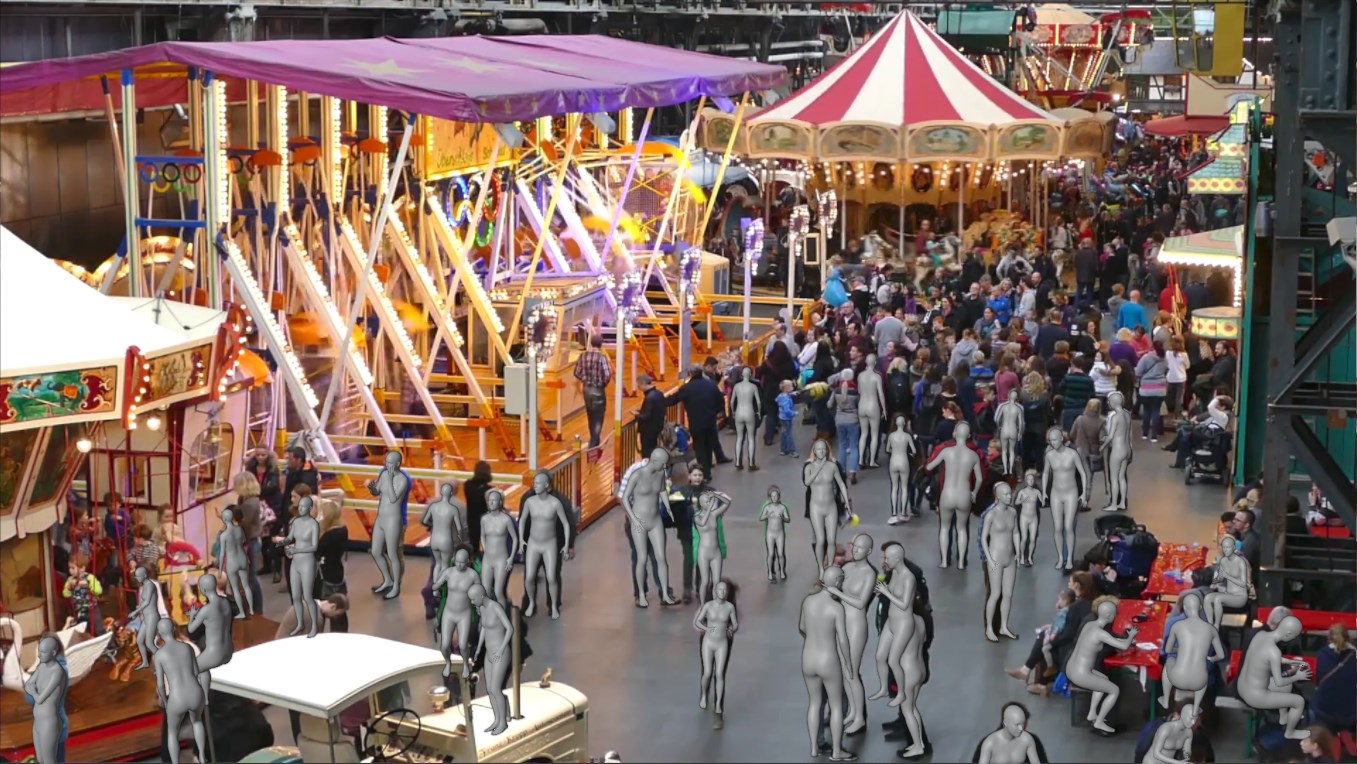}
        \hfill\includegraphics[width=.495\linewidth]{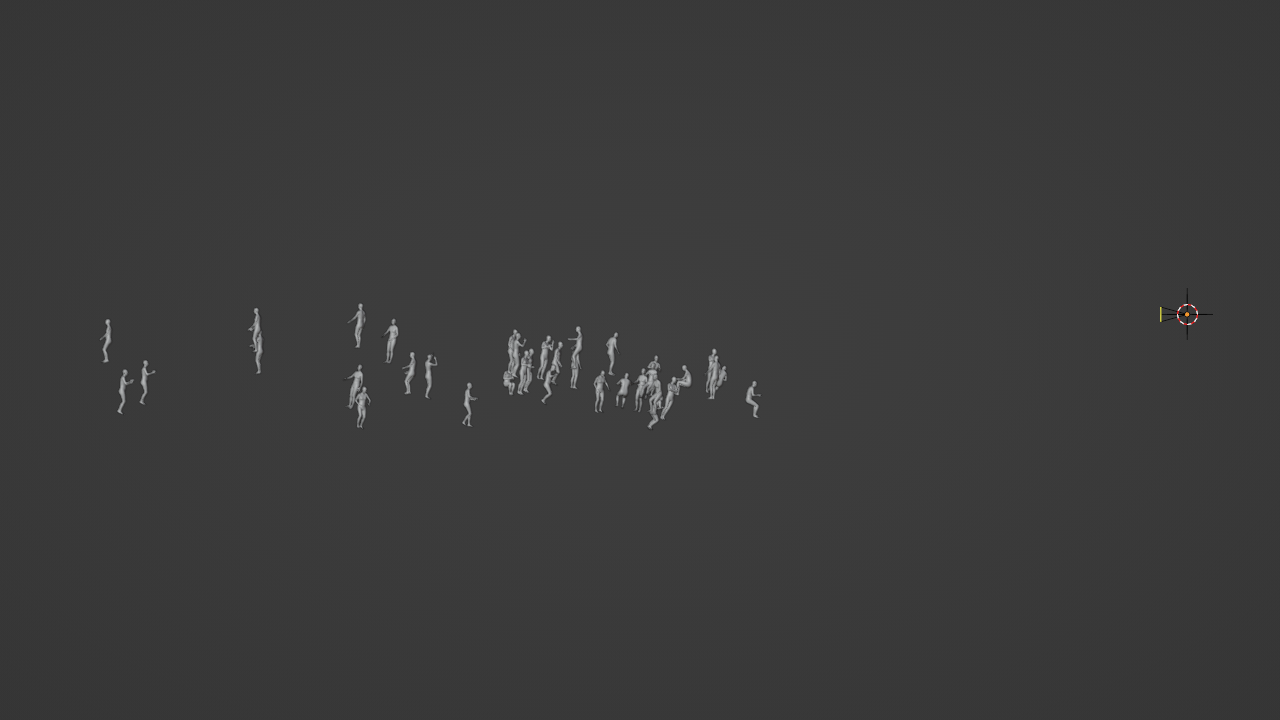}
            \caption{CLIFF \cite{li2022cliff}}
    \end{subfigure}

    \begin{subfigure}[h]{\linewidth}
        \includegraphics[width=0.495\linewidth]{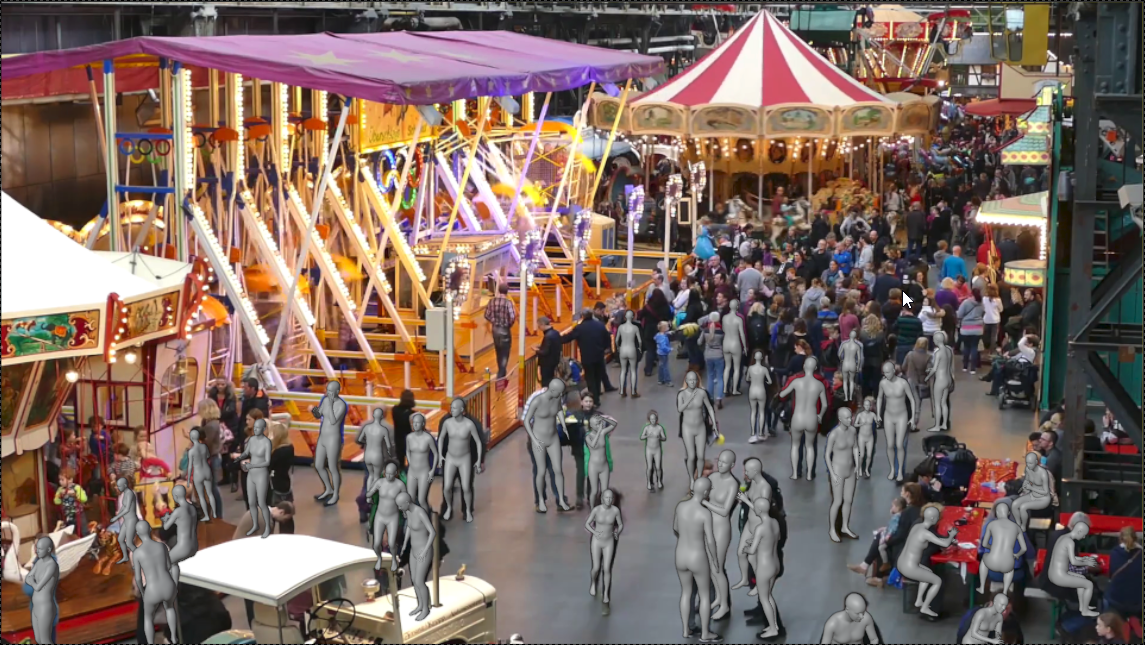}
        \hfill\includegraphics[width=.495\linewidth]{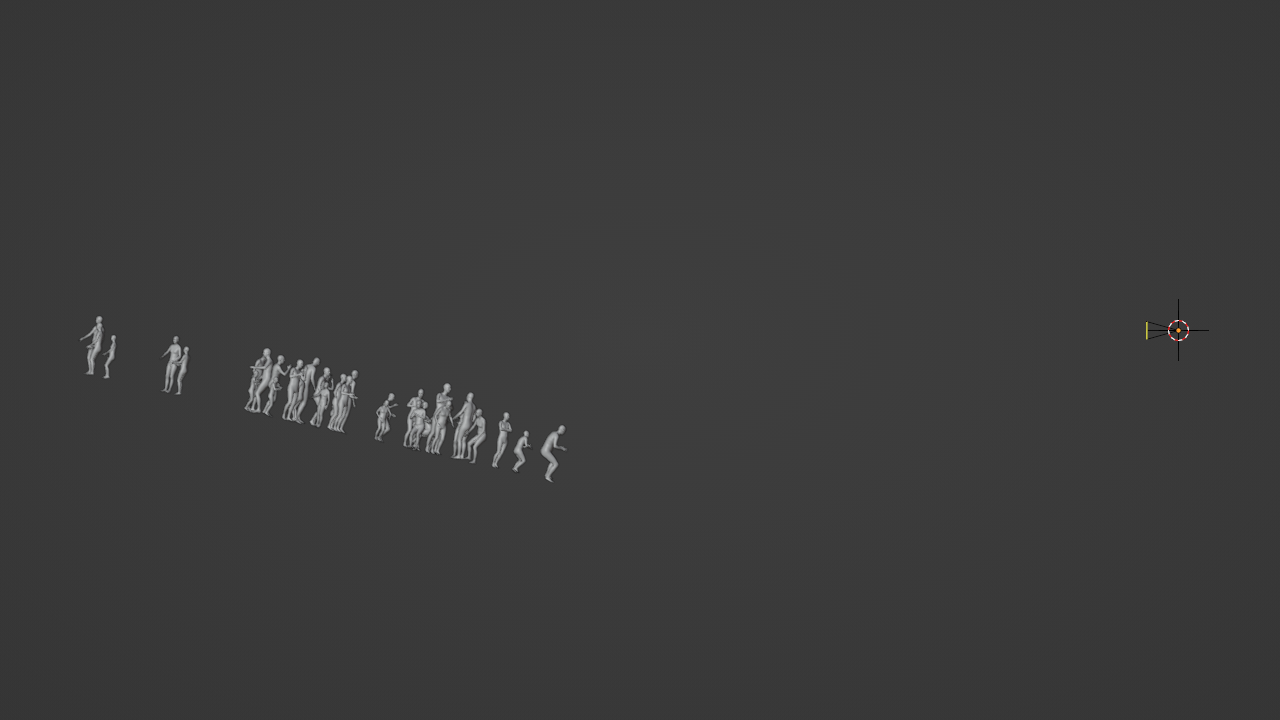}
            \caption{\textbf{RotAvat (ours)}}
    \end{subfigure}
    \caption{More comparison between BEV, SPEC, CLIFF and our proposed method, front and side views.}
    \label{fig_fair}
\end{figure}

\begin{figure}[h]
    \begin{subfigure}[h]{\linewidth}
        \includegraphics[width=0.473\linewidth]{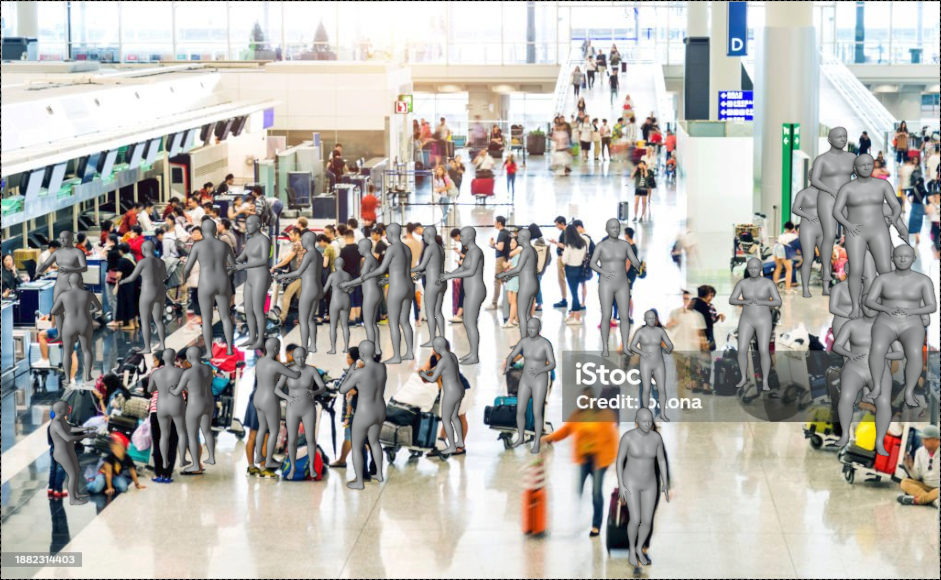}
        \hfill\includegraphics[width=.517\linewidth]{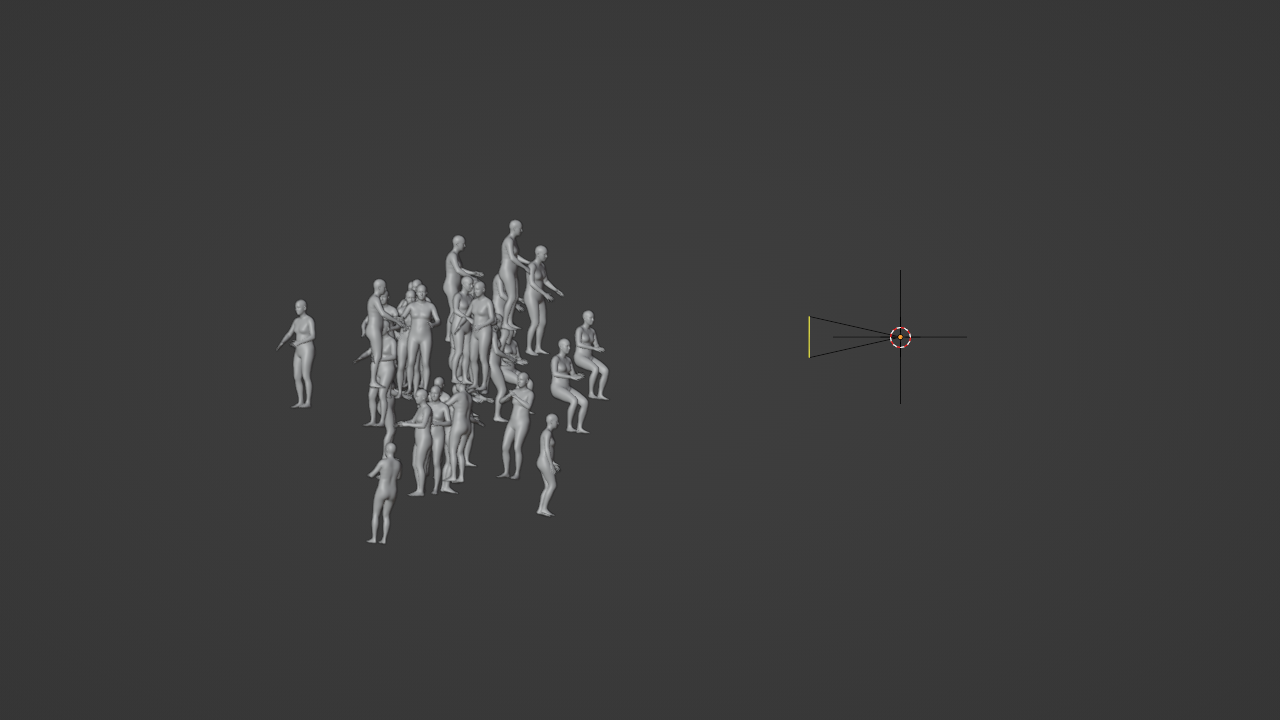}
            \caption{BEV \cite{BEV}
            }
    \end{subfigure}
    \begin{subfigure}[h]{\linewidth}
        \includegraphics[width=0.473\linewidth]{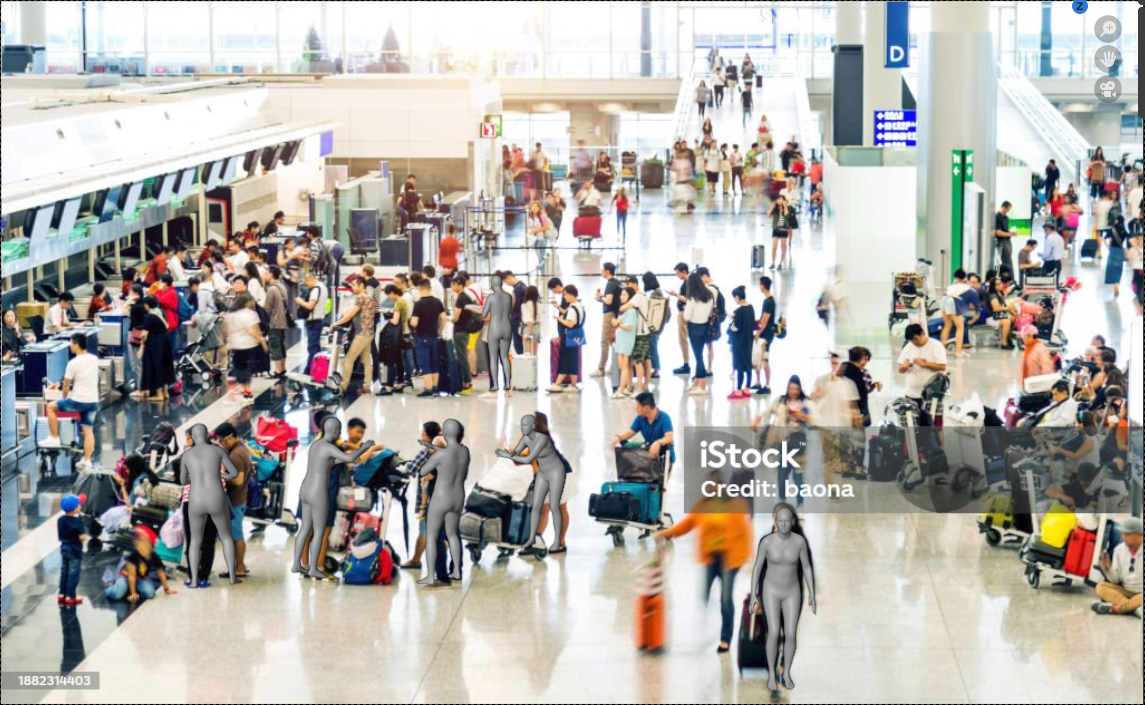}
        \hfill\includegraphics[width=.517\linewidth]{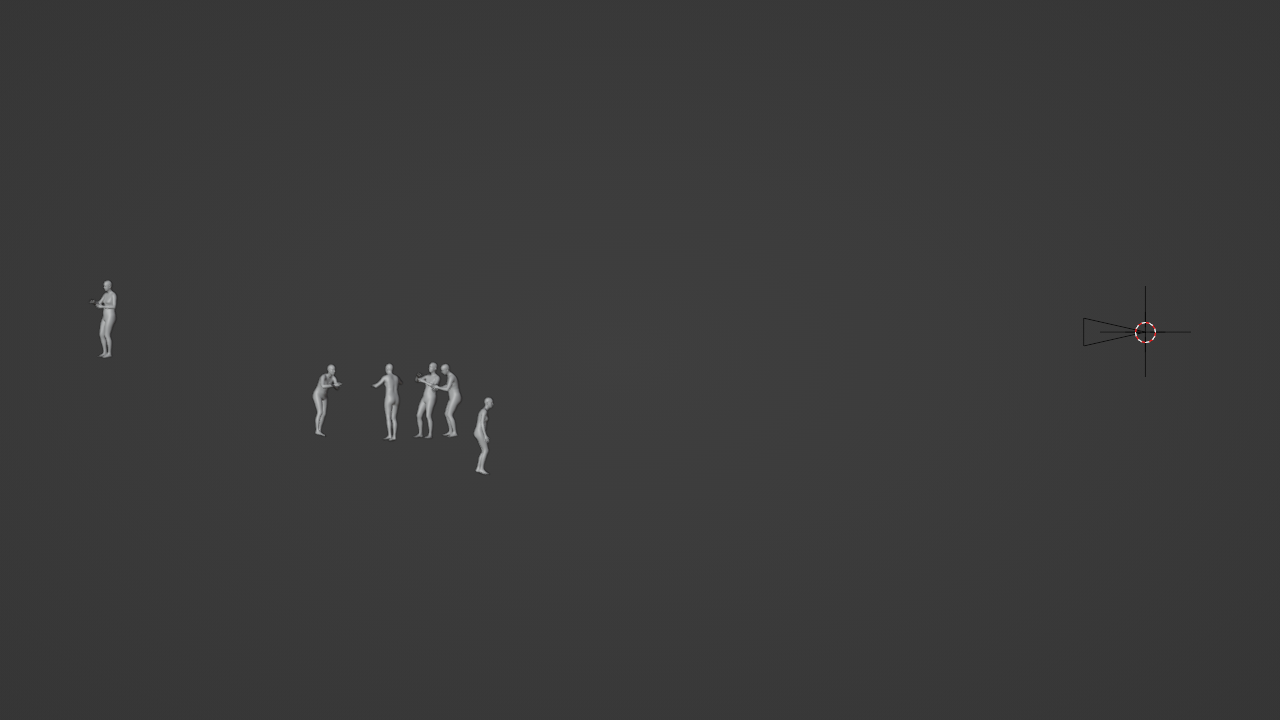}
            \caption{SPEC \cite{SPEC:ICCV:2021}}
    \end{subfigure}

    \begin{subfigure}[h]{\linewidth}
        \includegraphics[width=0.473\linewidth]{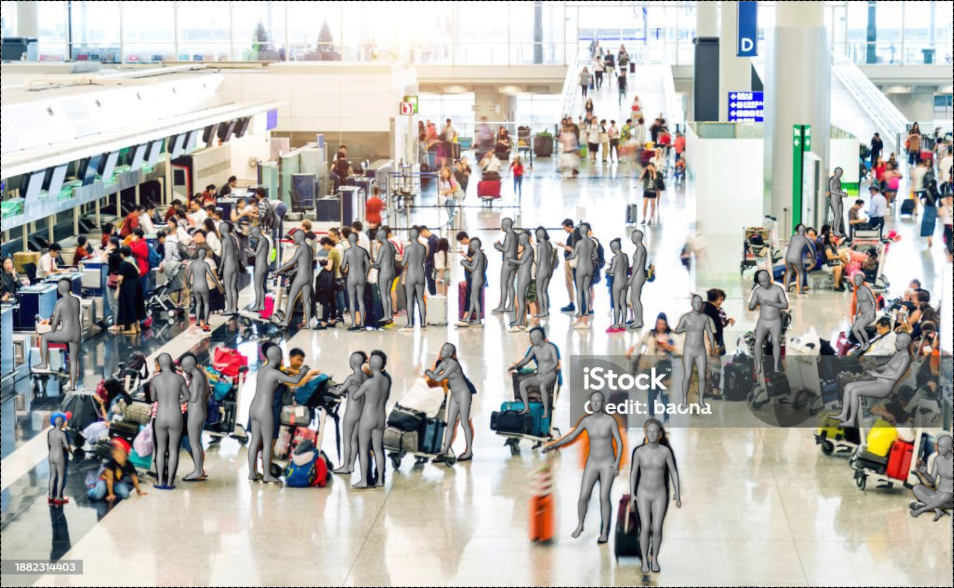}
        \hfill\includegraphics[width=.517\linewidth]{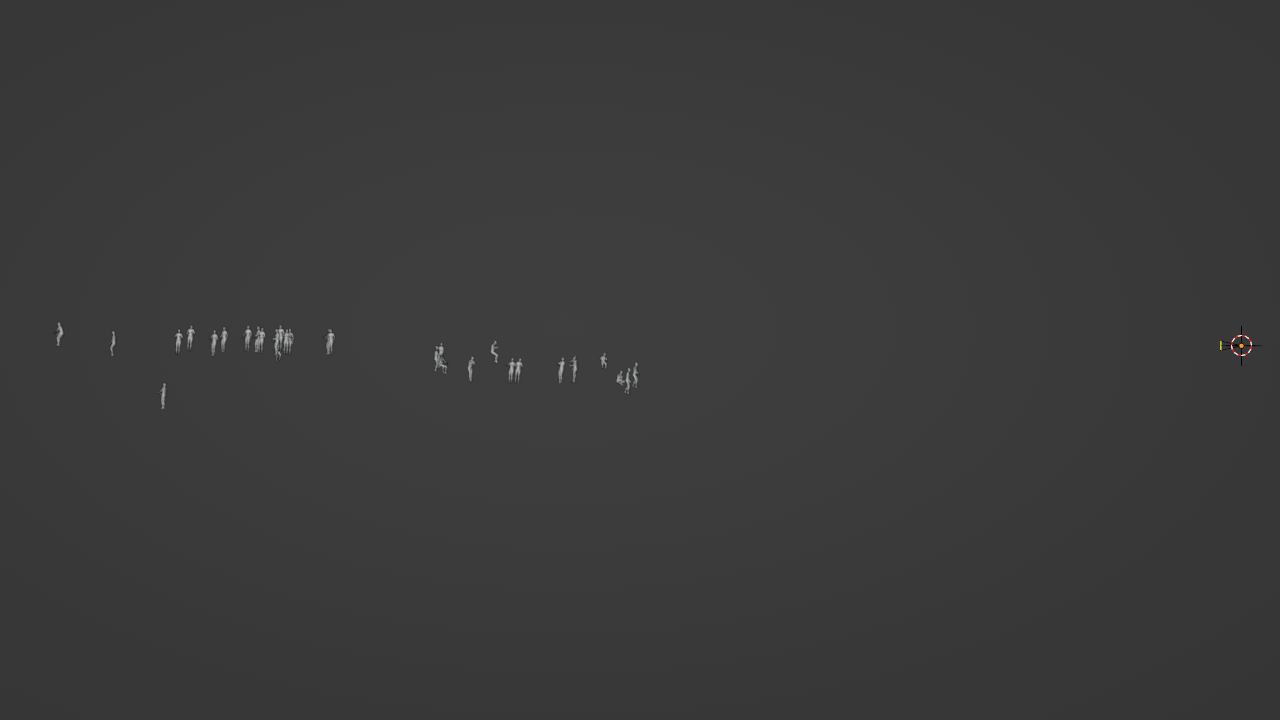}
            \caption{CLIFF \cite{li2022cliff}}
    \end{subfigure}

    \begin{subfigure}[h]{\linewidth}
        \includegraphics[width=0.473\linewidth]{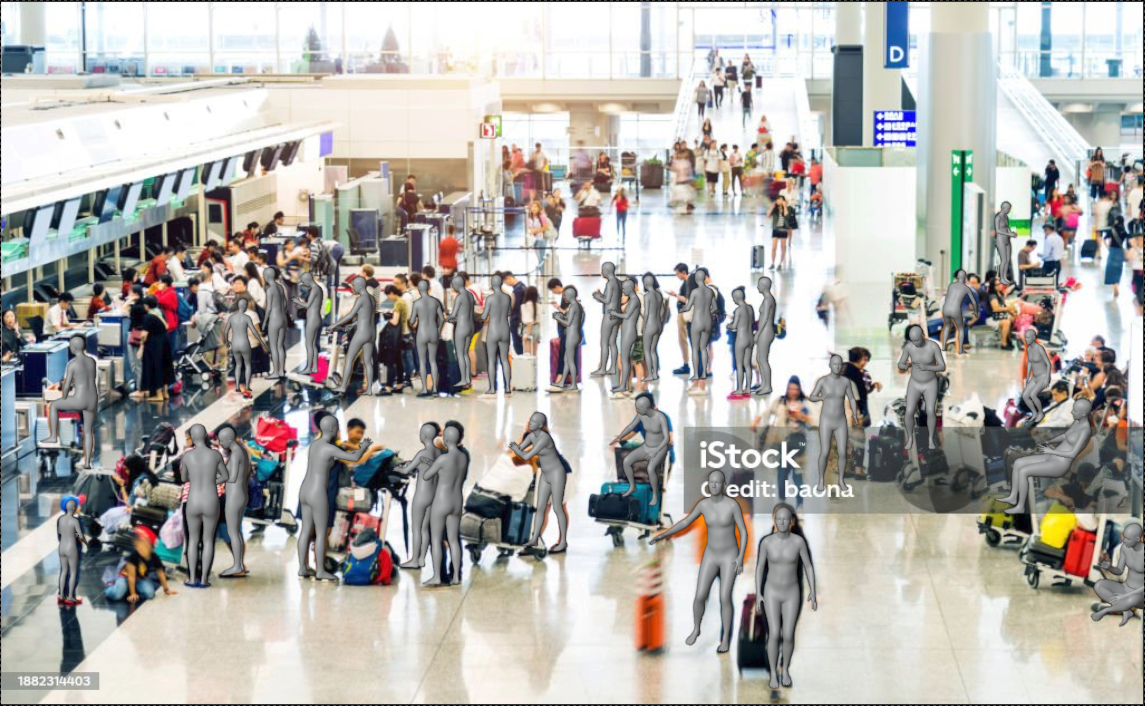}
        \hfill\includegraphics[width=.517\linewidth]{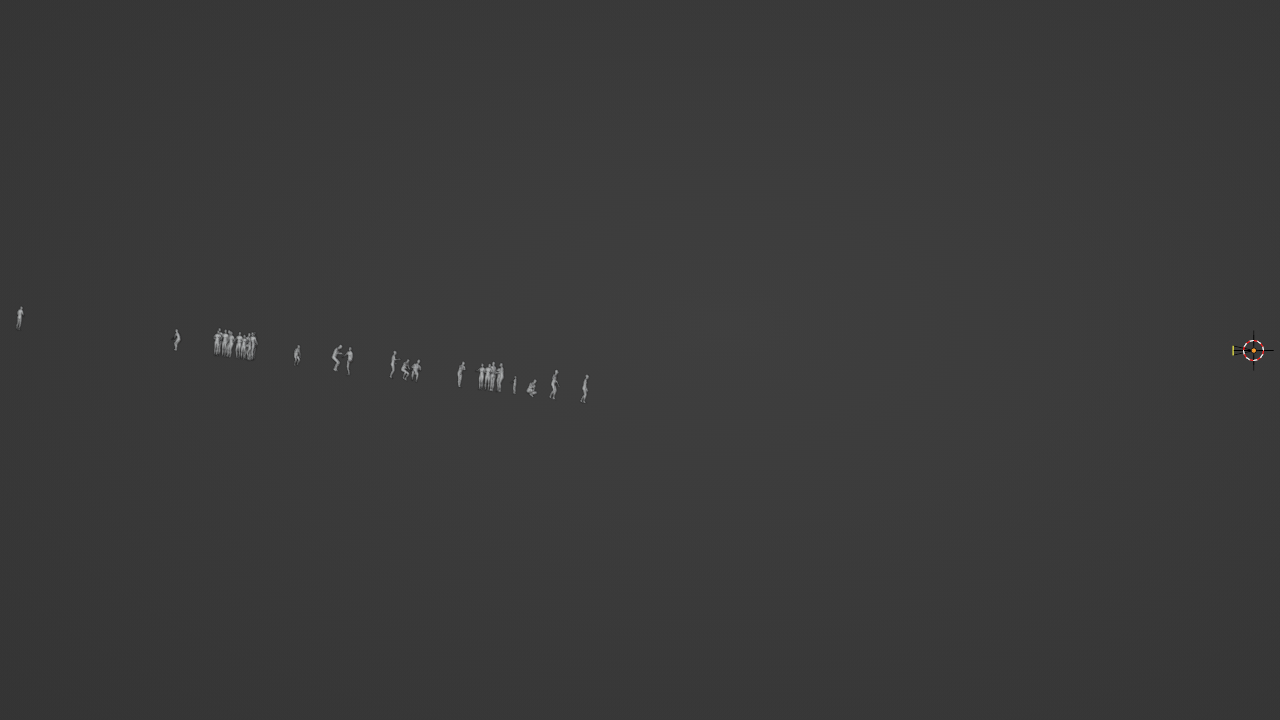}
            \caption{\textbf{RotAvat (ours)}}
    \end{subfigure}
    \caption{More comparison between BEV, SPEC, CLIFF and our proposed method, front and side views.}
    \label{fig_airport}
\end{figure}

\begin{figure}[h]
    \begin{subfigure}[h]{\linewidth}
        \includegraphics[width=0.453\linewidth]{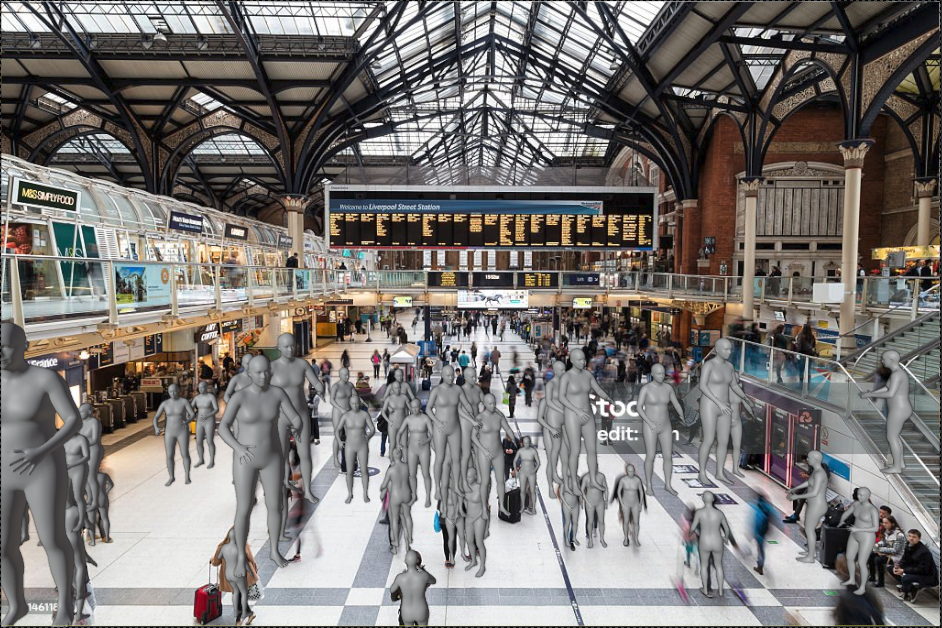}
        \hfill\includegraphics[width=.537\linewidth]{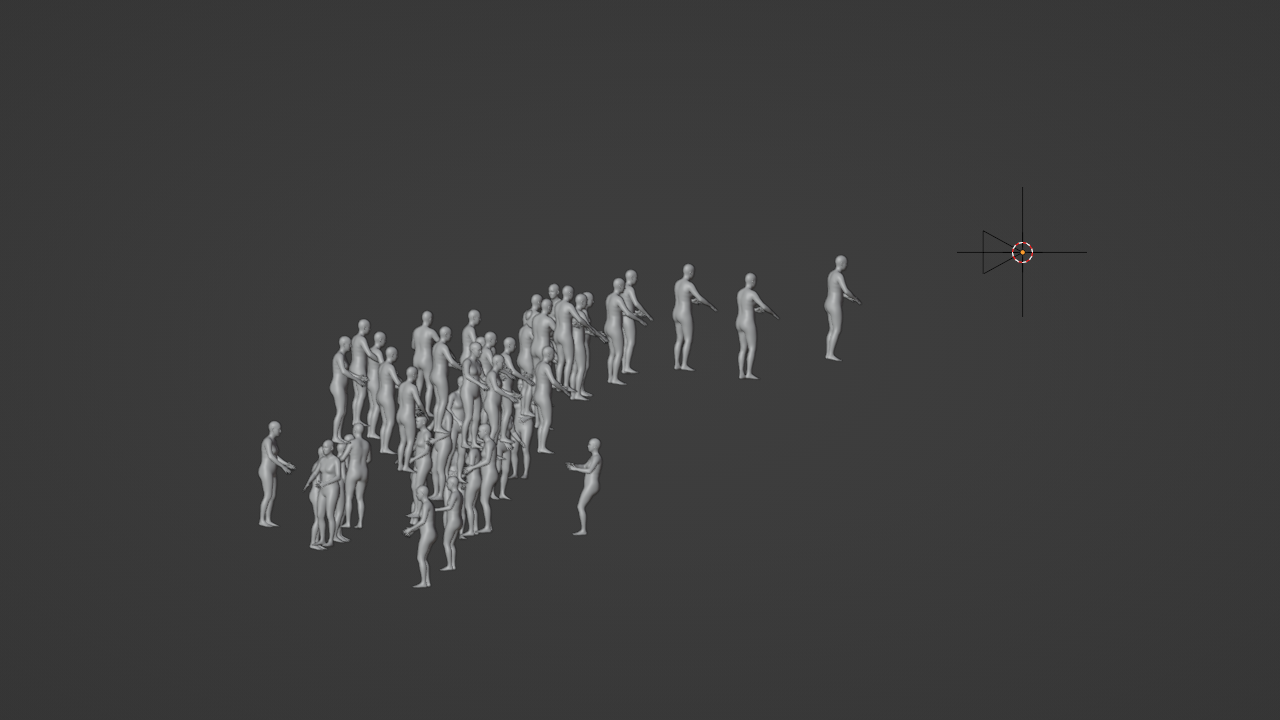}
            \caption{BEV \cite{BEV}
            }
    \end{subfigure}
    \begin{subfigure}[h]{\linewidth}
        \includegraphics[width=0.453\linewidth]{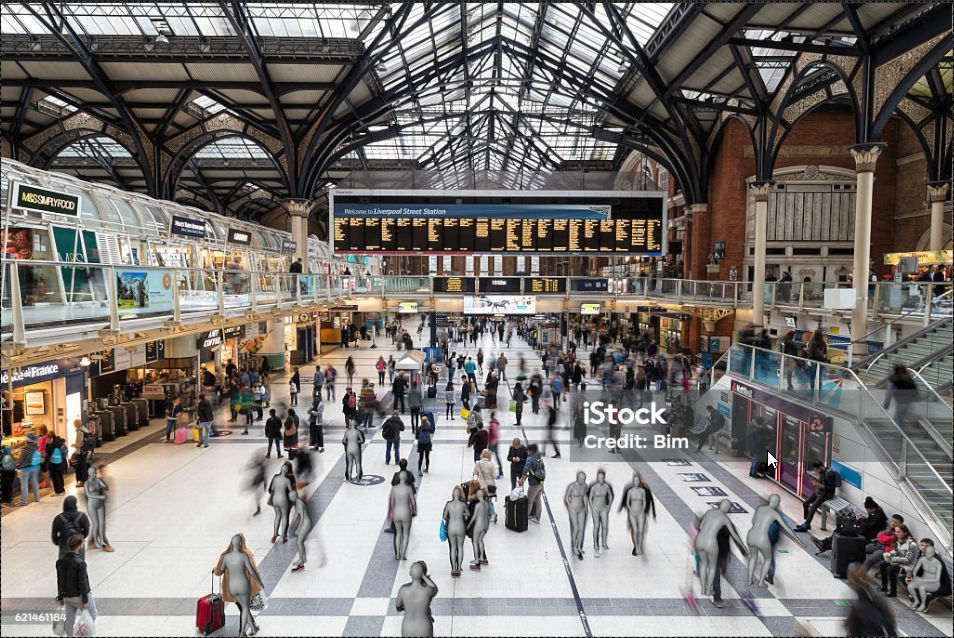}
        \hfill\includegraphics[width=.537\linewidth]{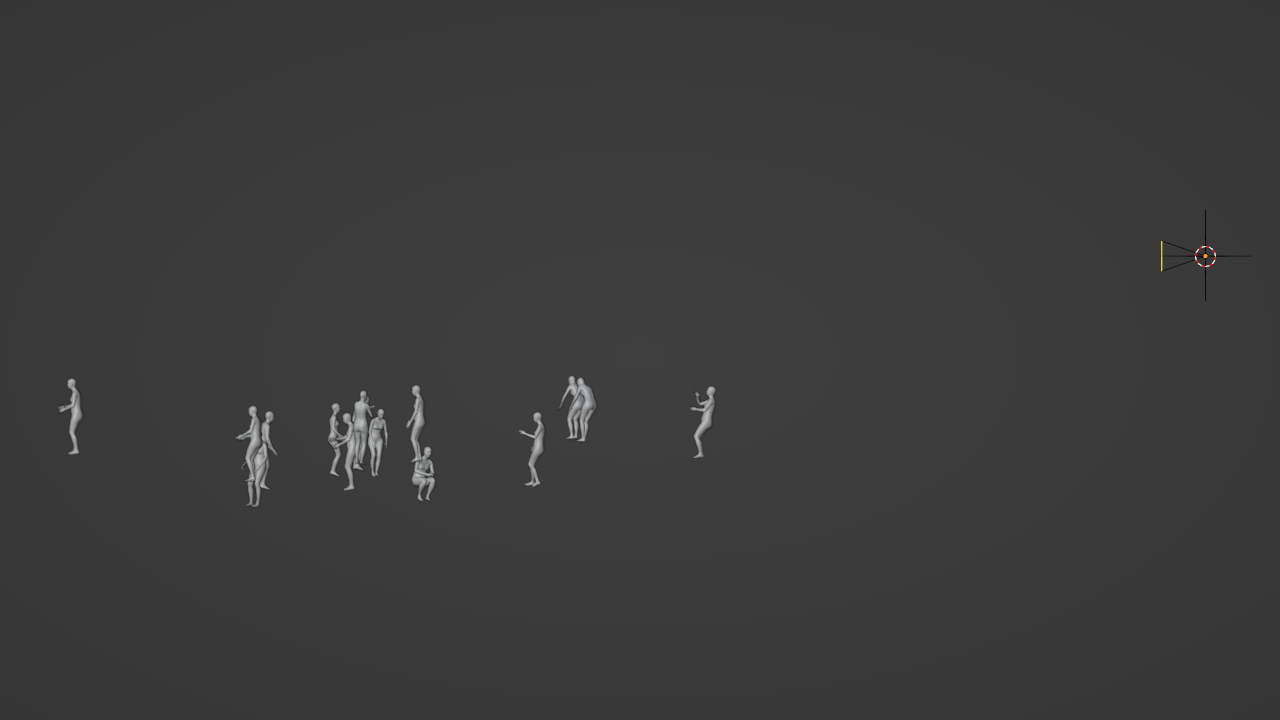}
            \caption{SPEC \cite{SPEC:ICCV:2021}}
    \end{subfigure}

    \begin{subfigure}[h]{\linewidth}
        \includegraphics[width=0.453\linewidth]{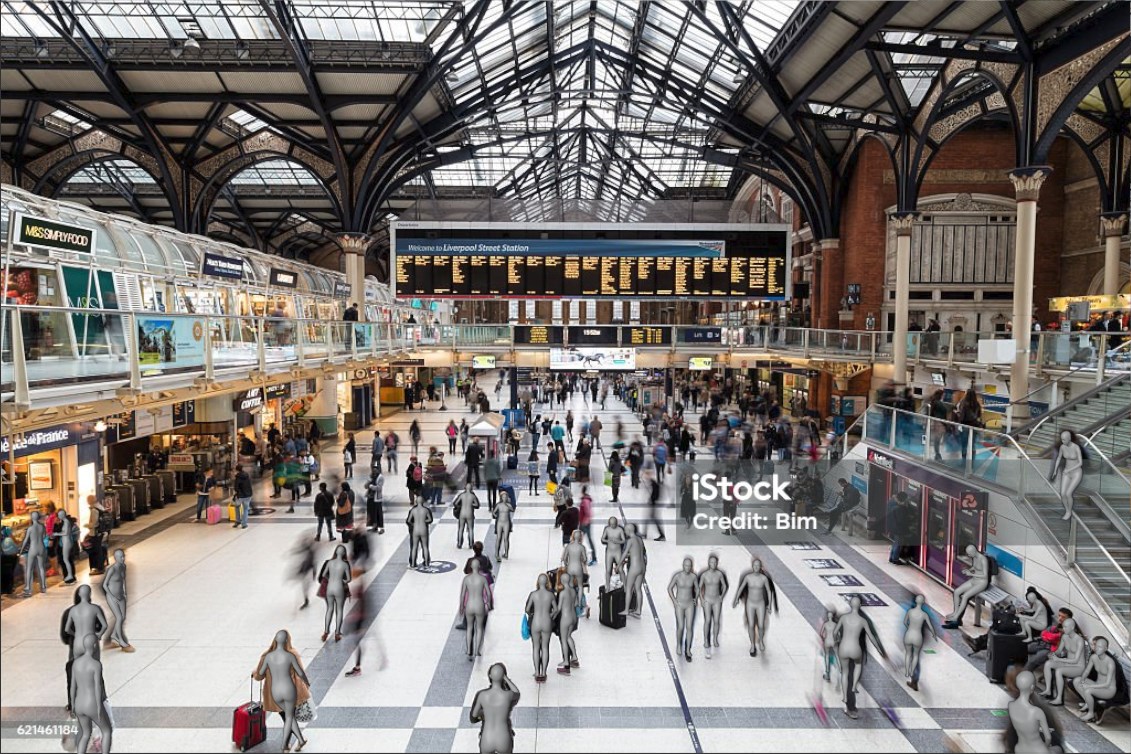}
        \hfill\includegraphics[width=.537\linewidth]{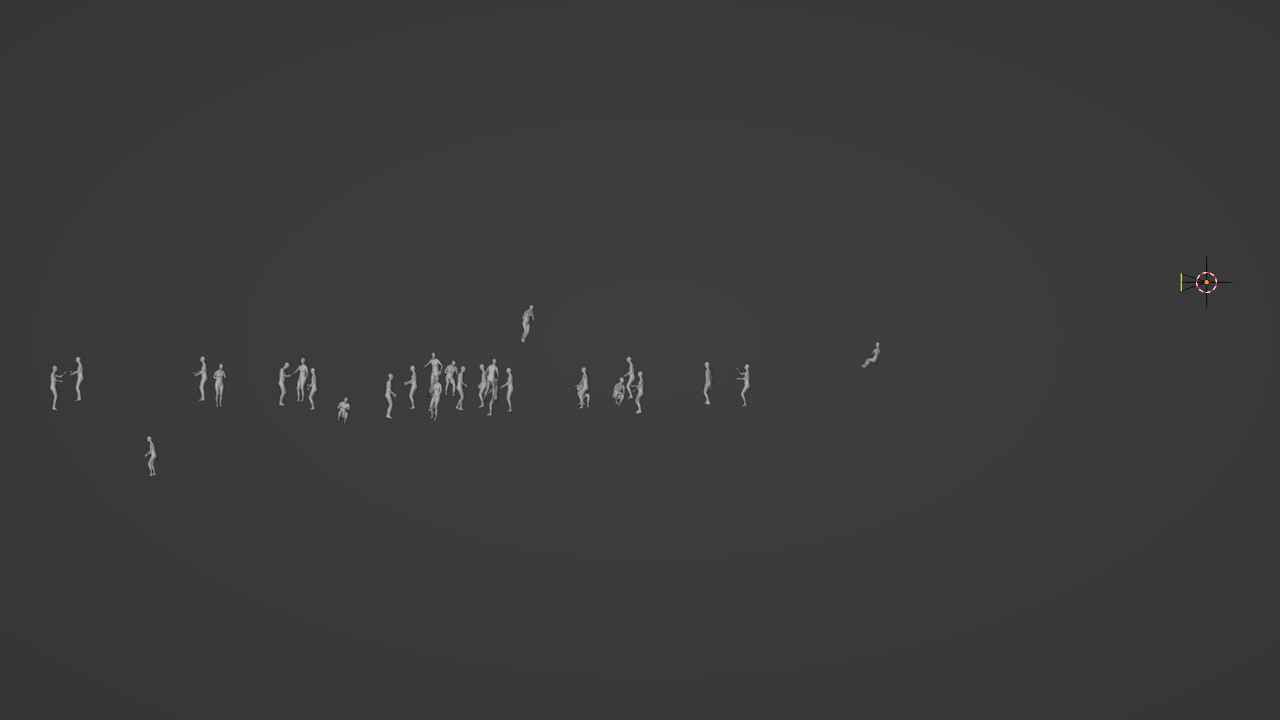}
            \caption{CLIFF \cite{li2022cliff}}
    \end{subfigure}

    \begin{subfigure}[h]{\linewidth}
        \includegraphics[width=0.453\linewidth]{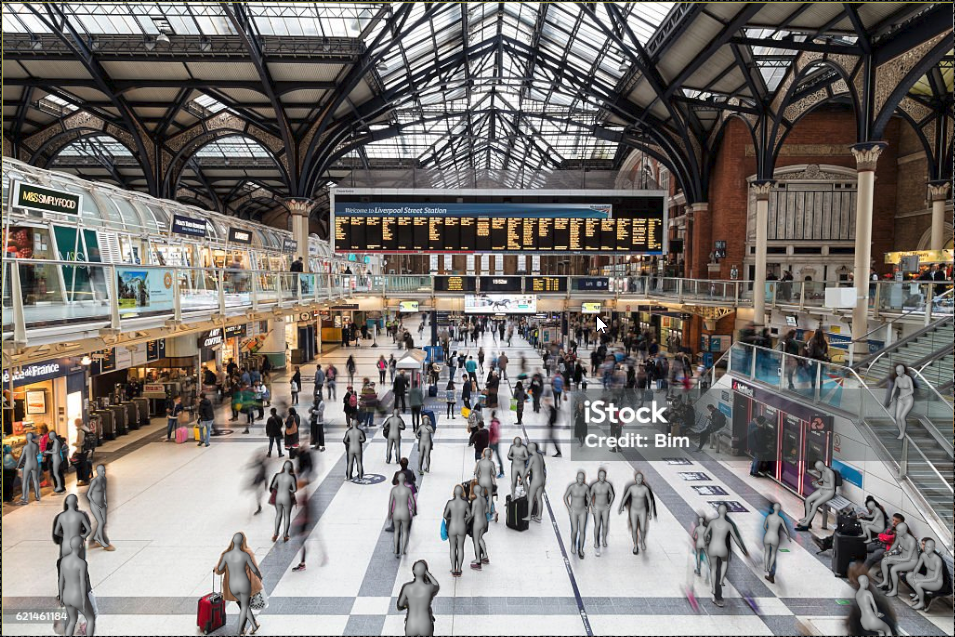}
        \hfill\includegraphics[width=.537\linewidth]{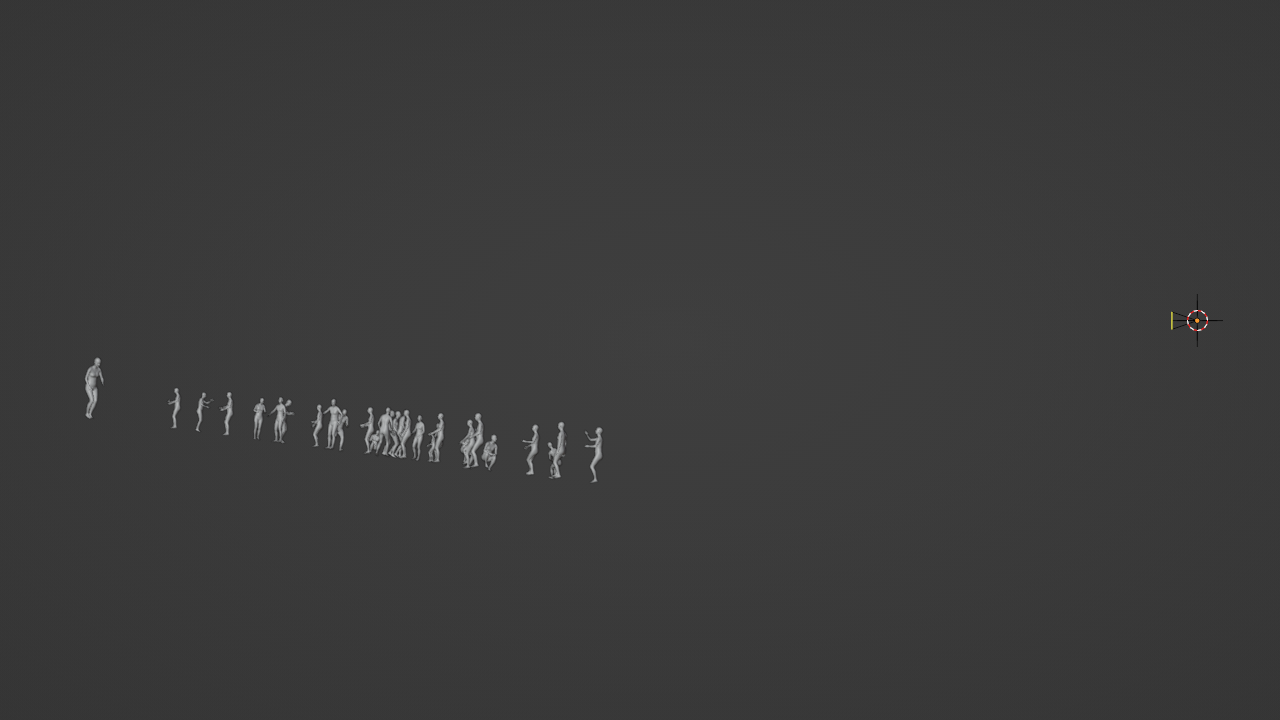}
            \caption{\textbf{RotAvat (ours)}}
    \end{subfigure}
    \caption{More comparison between BEV, SPEC, CLIFF and our proposed method, front and side views.}
    \label{fig_station}
\end{figure}

As we already mentioned, state-of-the-art methods may not be adaptable to unexpected variations encountered in in-the-wild scenarios. As a result, they may struggle in effectively addressing the W-MPJPE and W-PVE metrics.
In this section, we first confirm this using qualitative results in some video surveillance scenarios (which, let's recall, are the focus of our attention here). We compare those results with our approach, called \emph{RotAvat} (which stands for rotating avatars). Finally, we end this section with a detailed explanation of our approach.

\subsection{Qualitative comparison}
We compare \emph{RotAvat} with the main approaches available, which are the state-of-the-art methods for the HPS task (as indicated in our literature review): BEV \cite{BEV}, SPEC \cite{SPEC:ICCV:2021} and CLIFF \cite{li2022cliff}. The figures~\ref{fig_crowd_indoor},\ref{fig_fair},\ref{fig_airport},\ref{fig_station} provide a visual illustration of the deficiencies inherent in current approaches. First, we can mention the usual difficulties encountered by existing methods in accurately estimating poses, particularly in crowded scenes or cases of partial occlusion. This is already apparent in the rendering of the meshes from the front view. These minor errors in pose and shape are not the central aspect we wish to highlight, but are clearly those taken into account by the standard PA-MPJPE metric.
On the other hand, the camera perspective has a considerable influence on the results, revealing a lack of global consistency between the predicted meshes.
This is evident when the scene is rendered from a side view rather than the conventional front view (where there's nothing to indicate that the model has made such errors).
Indeed, in competing methods, rendered avatars often show poor positioning with respect to the ground, with noticeable problems of body rotation. Pedestrians tend to align themselves parallel to the camera rather than exhibiting the expected 90-degree rotation with respect to the ground. In our opinion, this indicates at least a lack of diversity of viewpoints in the training data set.

\subsection{Our approach} We now explain our RotAvat approach. 
We saw that despite pedestrians walking on flat ground, the 3D meshes predicted by state-of-the-art methods are not appropriately aligned with the ground level. To address this, we introduce two assumptions regarding the scene: firstly, that pedestrians are walking on flat ground, and secondly, that they are standing upright. 
These assumptions are often valid in camera surveillance scenarios. Our approach leverages these assumptions in two steps: the first is a simple auto-calibration of the camera based on the predicted foot and head keypoints. The second is the RotAvat step itself, which consists of transforming each 3D mesh by rotation, translation and scaling, in order to align and straighten them on the ground. The challenge is to manipulate the meshes while ensuring that this adjustment does not alter the good results observed from the camera's point of view (the front views in our figures).

\begin{figure}[h]
\includegraphics[width=\linewidth]{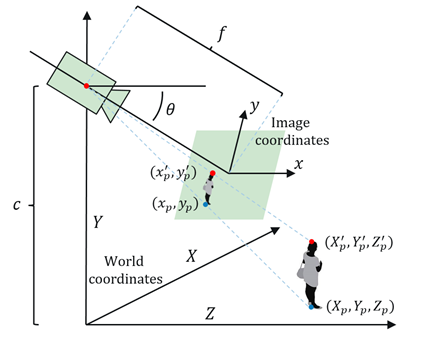}
    \caption{An example of a pedestrian with the foot-head point pair, with an example of camera calibration parameters we want to regress. For this choice of calibration parameters, the projection matrix of the camera is defined as $P=\begin{pmatrix}f & 0 & 0 & 0\\ 0 & f\cos(\theta) & -f\sin(\theta) & -fc\cos(\theta) \\ 0 &\sin(\theta)&\cos(\theta)&-c\sin(\theta)
    \\ 0&0&0&1
    \end{pmatrix}$.}
    \label{fig_calib}
\end{figure}

\paragraph{Auto-calibration}
It's important to emphasize that this module exclusively utilizes 2D input, precisely because it's considered robust compared to 3D, which requires improvement.
The auto-calibration phase begins by extracting the pairs of foot-head\footnote{$(x_p,y_p)$ corresponds to the average between the two points of the feet.} 2D points $(x_p,y_p),(x'_p,y'_p)$ for each pedestrian $p\in \mathcal{M}$ in the image ($\mathcal{M}$ being the set of predicted meshes).
The goal is to regress the camera calibration parameters from this set of foot-head pairs
(see Figure~\ref{fig_calib}). For each $p\in  \mathcal{M}$, given the projection matrix $P$ of the camera, a 3D foot-head pair is described by 6 degrees of freedom: 3 coordinates $X_p,Y_p,Z_p$   for the foot point, and 3 coordinates $X_p',Y_p',Z_p'$ for the head point. We assume that a pedestrian stands vertically to the ground and has a height of about 170 cm, so that we have $(Y_p,Y_p')=(0,170)$, and $(X_p,Z_p)=(X_p',Z_p')$. 
This means that only 2 degrees of freedom remain. 
More precisely, the 2D projection on the image of a 3D point $(X,Y,Z)$ is obtained as $(wx,wy,w,1) = P(X,Y,Z,1)$. Since $Y\in\{0,170\}$ is known, we have 
that $w=\frac{Y-P^{-1}_{1,3}}{P^{-1}_{1,:}(x,y,1,0)}$ is known.
We thus get a system of 4 equations, with 2 unknowns $(X_p,Z_p)$:
\begin{align}
    (x_p,y_p) &= \frac{1}{w}P_{:2,:}(X_p,0,Z_p,1),\label{1steq}\\ (x_p',y_p') &= \frac{1}{w'}P_{:2,:}(X_p,170,Z_p,1)\label{2ndeq}
\end{align}
There are several ways to simplify them into two equations containing no more unknowns, and we have chosen to first determine $X_p,Z_p$ using~\eqref{1steq}, and them re-inject them into the RHS of~\eqref{2ndeq} to get estimates $\hat x'_p, \hat y'_p$, function of only $P,x_p,y_p$.
The 2 equations $(x_p',y_p')=(\hat x'_p, \hat y'_p)$ are not exactly satisfied for each pedestrian $p\in \mathcal{M}$, so we can aggregate the errors over all $p\in \mathcal{M}$ and use the associated mean squared error to regress the calibration matrix $P$:
\[\text{MSE}(P)=\sum_{p\in \mathcal{M}}(x_p'-\hat x'_p)^2 + (y_p'-\hat y'_p)^2.\]
Depending on how the matrix $P$ is parameterized, we can use different optimization technique to minimize $\text{MSE}(P)$. We used only 3 parameters for $P$, namely the focal length $f$ of the camera,
the transverse tilt angle $\theta$ of the camera (called pitch) and
the camera height $c$. Thus, grid search is particularly adapted (plus, we know approximately the range of the 3 parameters). More precisely, we 
used the following ranges, with 50 bins for each of the 3 intervals (this gives 125000 values of $\text{MSE}(f,\theta,c)$ to check), where $H$ is the height of the image.
 \[f\in [0.1H,6H], \theta \in [-\pi/4,\pi/2], c\in [50,4000].\]

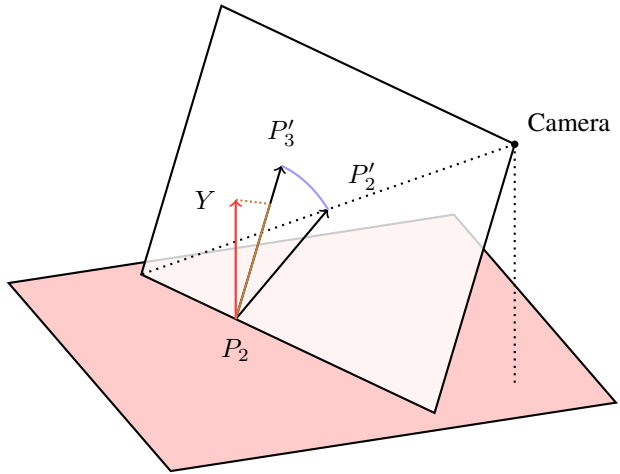
\begin{figure}[h]
\tdplotsetmaincoords{65}{70}
\begin{tikzpicture}[scale=0.35,tdplot_main_coords,bullet/.style={circle,inner
sep=1pt,fill=black,fill opacity=1}]

\coordinate (Camera) at (0,0,10);
\coordinate (P'2) at (-7,-5,5);
\coordinate (P'1) at (-9,-8,0);

\def\alpha{0.37818779029489047*5};
\def\beta{6.164414002968976/6.164414002968976};

\coordinate (Z) at ($(P'1) + (0,0,5)$);

\coordinate (n) at (-0.34380708*\alpha,  0.85951771*\alpha, -0.37818779*\alpha);
\coordinate (u) at (0.92847669, 0.37139068, 0.);

\coordinate (Pz) at ($ (n) + (0,0,5) + (P'1)$);

\begin{scope}[canvas is xy plane at z=0]
\draw[thick,fill=red!20] (-15,-15) rectangle (3,3);
\end{scope}
 %\begin{scope}[canvas is xz plane at y=1.5]
  %\draw[thick,fill=white,fill opacity=0.7,nodes={opacity=1}]
   %(0,0) node[bullet,label=below:{$\mathsf{x}\beta$}] {}
   %-- (2,0) node[bullet,label=below:{$\mathsf{x}\hat\beta$}] {}
   %-- (2,3) node[bullet,label=above right:{\text{Camera}}] {} -- cycle;
   %\draw (1.75,0) -- (1.75,0.25) -- (2,0.25);
 %\end{scope}
 %\draw (-1,0,0)--(0,0,1);
 %\draw (-1,0,0)--(0,0,0)--(0,0,1);
\draw[thick, dotted](0,0,0)--(0,0,10);
\draw[thick,fill=white,fill opacity=.8,nodes={opacity=1}]
  (0,0,10) node[bullet,label=above right:{\text{Camera}}] {}
    -- (-15.51724138,  -6.20689655,  10.) node[label=above:{}] {}
    -- (-14,-10,0) node[label=below:{}] {}
    -- (P'1) node[label=below:{$P_2$}] {}
    -- (1.51724138, -3.79310345,  0) node[label=below:{}] {}
    -- cycle;

\draw [thick, dotted]  (Camera) -- (P'2) node[label=above right:{$P_2'$}]{} -- (-14,-10,0);
\draw [->, thick] (P'1)--(P'2);
\draw [->, thick] (P'1)--(-9.86582535, -5.83543662,  5.70657619)node[label=above:{$P_3'$}]{};

% draw rigth angle on The Z axis at P'1
\draw [thick, ->, draw=red!80] (P'1)--(Z)node[label=left:{$Y$}]{};
% draw projection of Z on Plan
\draw [thick,-, draw=brown] (-9,-8,0)--(Pz);

\draw[thick]
    -- (P'1)
    -- ($(P'1) - (-0.18569534,  0.46423835,  0)$)
    -- ($(P'1) - (-0.18569534,  0.46423835,  -0.5)$)
    -- ($(P'1) + (0,0,0.5)$)
    -- cycle;

% draw dashed line of the projection
\draw [thick,densely dotted, draw=brown] (Z)--(Pz);

\coordinate (P'1toP''2) at (-0.14045542*0.5,  0.35113855*0.5,  0.9257289*0.5);
\coordinate (ZtoPz) at (-0.34380708*0.5,  0.85951771*0.5, -0.37818779*0.5);

% mean of P''2 and P'2 0.56708732, 2.58228169, 5.3532881

% draw rigth angle
\draw[thick]
    -- (0,0,0)
    -- (0,-0.5,0)
    -- (0,-0.5,0.5)
    -- (0,0,0.5)
    -- cycle;

% draw rigth angle
\draw[thick]
    -- (Pz)
    -- ($(Pz) - (P'1toP''2)$)
    -- ($(Pz) - (P'1toP''2) - (ZtoPz)$)
    -- ($(Pz) - (ZtoPz)$)
    -- cycle;
    
% draw angle
\draw [blue!40,thick,domain=90:119] plot 
({-9 - cos(\x) * 5.72351471 * \beta + sin(\x) * -0.86582536 * \beta},
{-8  - cos(\x) * 2.28940589 * \beta + sin(\x) * 2.16456334 * \beta}, 
{0   - cos(\x) * 0          + sin(\x) * 5.70657621 * \beta});
    
\end{tikzpicture} 
        \caption{The rotation step in RotAvat.}
    \label{fig_rotavat1}
\end{figure}

\begin{figure}[h]
\tdplotsetmaincoords{65}{70}
\begin{tikzpicture}[scale=0.35,tdplot_main_coords,bullet/.style={circle,inner
sep=1pt,fill=black,fill opacity=1}]

\coordinate (Camera) at (0,0,10);
\coordinate (P'2) at (-7,-5,5);
\coordinate (P''2) at (-9.86582535, -5.83543662,  5.70657619);
\coordinate (P'''2) at (-9.48919132, -6.77702169,  3.22421555);
\coordinate (P'1) at (-9,-8,0);

\def\alpha{0.37818779029489047*5};
\def\beta{6.164414002968976/6.164414002968976};

\coordinate (Z) at ($(P'1) + (0,0,5)$);

\coordinate (n) at (-0.34380708*\alpha,  0.85951771*\alpha, -0.37818779*\alpha);
\coordinate (u) at (0.92847669, 0.37139068, 0.);

\coordinate (Pz) at ($ (n) + (0,0,5) + (P'1)$);

\begin{scope}[canvas is xy plane at z=0]
\draw[thick,fill=red!20] (-15,-15) rectangle (3,3);
\end{scope}
 %\begin{scope}[canvas is xz plane at y=1.5]
  %\draw[thick,fill=white,fill opacity=0.7,nodes={opacity=1}]
   %(0,0) node[bullet,label=below:{$\mathsf{x}\beta$}] {}
   %-- (2,0) node[bullet,label=below:{$\mathsf{x}\hat\beta$}] {}
   %-- (2,3) node[bullet,label=above right:{\text{Camera}}] {} -- cycle;
   %\draw (1.75,0) -- (1.75,0.25) -- (2,0.25);
 %\end{scope}
 %\draw (-1,0,0)--(0,0,1);
 %\draw (-1,0,0)--(0,0,0)--(0,0,1);
\draw[thick, dotted](0,0,0)--(0,0,10);
\draw[thick,fill=white,fill opacity=.8,nodes={opacity=1}]
  (0,0,10) node[bullet,label=above right:{\text{Camera}}] {}
    -- (-15.51724138,  -6.20689655,  10.) node[label=above:{}] {}
    -- (-14,-10,0) node[label=below:{}] {}
    -- (P'1) node[label=below:{$P_2$}] {}
    -- (1.51724138, -3.79310345,  0) node[label=below:{}] {}
    -- cycle;

\draw [thick, dotted]  (Camera) -- (P'2) node[label=above right:{$P_2'$}]{} -- (-14,-10,0);
\draw [->, thick] (P'1)--(P'2);
\draw [dashed, ->, thick] (P'1)--(P''2)node[label=above:{$P_3'$}]{};
\draw [->, thick] (P'1)--(P'''2)node[label=left:{$P_4'$}]{};

% draw rigth angle on The Z axis at P'1
% \draw [thick, ->, draw=red!80] (P'1)--(Z)node[label=left:{Z}]{};
% draw projection of Z on Plan
% \draw [thick,-, draw=brown] (-9,-8,0)--(Pz);

% \draw[thick]
%     -- (P'1)
%     -- ($(P'1) - (-0.18569534,  0.46423835,  0)$)
%     -- ($(P'1) - (-0.18569534,  0.46423835,  -0.5)$)
%     -- ($(P'1) + (0,0,0.5)$)
%     -- cycle;

% draw rigth angle
\draw[thick]
    -- (0,0,0)
    -- (0,-0.5,0)
    -- (0,-0.5,0.5)
    -- (0,0,0.5)
    -- cycle;

\end{tikzpicture} 
        \caption{The last scaling step in RotAvat.}
    \label{fig_rotavat2}
\end{figure}
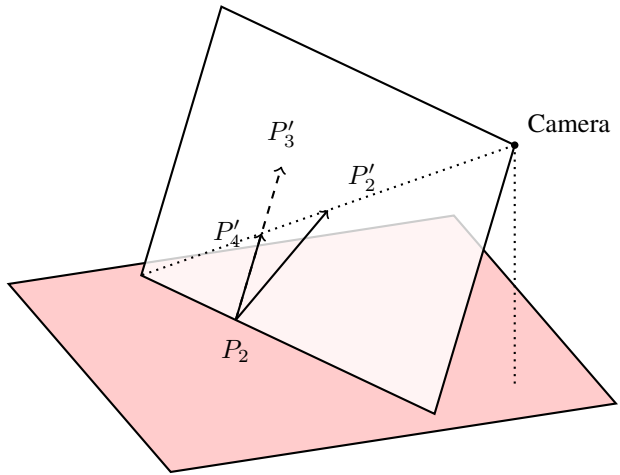

\paragraph{RotAvat}
From the previous calibration module, we now have the ground level available (a 2D plan in the 3D space). Our goal is to adjust meshes to align with the ground plane without compromising the results observed from the camera's perspective. Let's fix a given mesh $m\in \mathcal{M}$, and call $(P_1,P_1')$ its 3D foot-head pair. RotAvat is all based on applying three fundamental transformations: translating, rotating, and scaling of $m$. There are 3 phases: $(i)$ a scaling w.r.t. the camera center, $(ii)$ a rotation, and $(iii)$ a last scaling w.r.t. the foot point. These two scaling operations can be seen as scaling with respect to the mesh's center of gravity and then applying the appropriate translation.
\begin{itemize}
    \item[$(i)$] Note that we can always move away or bring the mesh $m$ closer to the camera without changing its aspect by applying a homothetic transformation taking the camera as the centric point. This allows us to set $m$ on the ground without changing anything from the camera point of view. Therefore, $(P_1,P_1')$ is transformed into $(P_2,P_2')$ such that $P_2$ is on the ground (the plan $Y=0$). However, $m$ may still be tilted relative to the ground (and may now be a bit scaled).
    \item[$(ii)$] One naive solution to get $(P_3,P_3')$ would be to adjust $m$ to compensate for the tilt by aligning it with the normal to the ground $Y$ (with some rotation of center $P_2$). Nevertheless, this would lead to significant alterations in the camera's point of view.
Instead of aligning the mesh directly based on the ground normal $Y$, we propose to align them based on the projection of $Y$ onto the plane containing the camera position, the mesh head position $P'_2$, and the middle foot position of the mesh $P_2$ (see the white plane in Figure~\ref{fig_rotavat1}). In practice, we observe that the mesh rotated in this way retains a satisfactory vertical appearance. We'll explain why this choice of rotation limits the final impact on rendering.
\item[$(iii)$] Finally, we perform one scaling with respect to the center $P_3=P_2$ as described in Figure~\ref{fig_rotavat2}. This is to put back the head point in the right position in the image.
\end{itemize}
The reason why this method doesn't change the rendered image very much is that the selected plane in phase $(ii)$ appears as a straight line from the camera's point of view. Consequently, the direction previously defined from $P_2$ to $P'_2$ is collinear with the rotated direction from $P_3$ to $P_3'$ in the camera's point of view. In other words, in the image, this rotation is akin to scaling if the object in question is thin enough (as is the case with pedestrians), and can thus be compensated for by the final scaling of phase $(iii)$.
All these steps therefore ensure that the foot point and head point before and after the transformation are positioned exactly at the same point in the camera's point of view.
  Note that the described method is independent from of the way in which the meshes are provided and can therefore be applied to any method. In our experiments, we chose the CLIFF \cite{li2022cliff} approach as input to RotAvat. 

\section{Conclusion}
\label{conclusion}
Our findings emphasize the importance of continuous innovation in addressing the evolving challenges of video surveillance. By introducing RotAvat, we offer a robust post-processing solution that enhances the accuracy of 3D human pose and shape estimation, particularly in the context of video surveillance with stationary cameras. Our approach effectively addresses the limitations of existing methods by ensuring better alignment and positioning of 3D meshes with respect to the ground plane, as demonstrated through qualitative comparisons. This work underscores the necessity of integrating real-world constraints into algorithmic design to achieve more reliable and practical outcomes in surveillance scenarios. Future research should explore the applicability of RotAvat to dynamic camera settings to fully leverage its potential in diverse surveillance environments.

%%%%%% REFERENCES
%\newpage
{
\small
\bibliographystyle{ieee_fullname}
\bibliography{ref}

\begin{thebibliography}{10}\itemsep=-1pt

\bibitem{black2023bedlam}
Michael~J Black, Priyanka Patel, Joachim Tesch, and Jinlong Yang.
\newblock Bedlam: A synthetic dataset of bodies exhibiting detailed lifelike animated motion.
\newblock In {\em Proceedings of the IEEE/CVF Conference on Computer Vision and Pattern Recognition}, pages 8726--8737, 2023.

\bibitem{SPEC:ICCV:2021}
Muhammed Kocabas, Chun-Hao~P. Huang, Joachim Tesch, Lea M\"uller, Otmar Hilliges, and Michael~J. Black.
\newblock {SPEC}: Seeing people in the wild with an estimated camera.
\newblock In {\em Proc. International Conference on Computer Vision (ICCV)}, pages 11035--11045, Oct. 2021.

\bibitem{li2023hybrik}
Jiefeng Li, Siyuan Bian, Chao Xu, Zhicun Chen, Lixin Yang, and Cewu Lu.
\newblock Hybrik-x: Hybrid analytical-neural inverse kinematics for whole-body mesh recovery.
\newblock {\em arXiv preprint arXiv:2304.05690}, 2023.

\bibitem{li2021hybrik}
Jiefeng Li, Chao Xu, Zhicun Chen, Siyuan Bian, Lixin Yang, and Cewu Lu.
\newblock Hybrik: A hybrid analytical-neural inverse kinematics solution for 3d human pose and shape estimation.
\newblock In {\em Proceedings of the IEEE/CVF Conference on Computer Vision and Pattern Recognition}, pages 3383--3393, 2021.

\bibitem{li2022cliff}
Zhihao Li, Jianzhuang Liu, Zhensong Zhang, Songcen Xu, and Youliang Yan.
\newblock Cliff: Carrying location information in full frames into human pose and shape estimation.
\newblock In {\em ECCV}, 2022.

\bibitem{ROMP}
Yu Sun, Qian Bao, Wu Liu, Yili Fu, Black Michael~J., and Tao Mei.
\newblock {Monocular, One-stage, Regression of Multiple 3D People}.
\newblock In {\em ICCV}, 2021.

\bibitem{TRACE}
Yu Sun, Qian Bao, Wu Liu, Tao Mei, and Michael~J. Black.
\newblock {TRACE: 5D Temporal Regression of Avatars with Dynamic Cameras in 3D Environments}.
\newblock In {\em CVPR}, 2023.

\bibitem{BEV}
Yu Sun, Wu Liu, Qian Bao, Yili Fu, Tao Mei, and Michael~J Black.
\newblock {Putting People in their Place: Monocular Regression of 3D People in Depth}.
\newblock In {\em CVPR}, 2022.

\bibitem{tripathi20233d}
Shashank Tripathi, Lea M{\"u}ller, Chun-Hao~P Huang, Omid Taheri, Michael~J Black, and Dimitrios Tzionas.
\newblock 3d human pose estimation via intuitive physics.
\newblock In {\em Proceedings of the IEEE/CVF Conference on Computer Vision and Pattern Recognition}, pages 4713--4725, 2023.

\bibitem{yi2023mime}
Hongwei Yi, Chun-Hao~P Huang, Shashank Tripathi, Lea Hering, Justus Thies, and Michael~J Black.
\newblock Mime: Human-aware 3d scene generation.
\newblock In {\em Proceedings of the IEEE/CVF Conference on Computer Vision and Pattern Recognition}, pages 12965--12976, 2023.

\bibitem{yuan2022glamr}
Ye Yuan, Umar Iqbal, Pavlo Molchanov, Kris Kitani, and Jan Kautz.
\newblock Glamr: Global occlusion-aware human mesh recovery with dynamic cameras.
\newblock In {\em Proceedings of the IEEE/CVF Conference on Computer Vision and Pattern Recognition (CVPR)}, 2022.

\bibitem{zhang2021pymaf}
Hongwen Zhang, Yating Tian, Xinchi Zhou, Wanli Ouyang, Yebin Liu, Limin Wang, and Zhenan Sun.
\newblock Pymaf: 3d human pose and shape regression with pyramidal mesh alignment feedback loop.
\newblock In {\em Proceedings of the IEEE/CVF International Conference on Computer Vision}, pages 11446--11456, 2021.

\end{thebibliography}
}

\onecolumn
%{\input{sections/7_appendix}}

\end{document}